%% file: main.tex
\documentclass[final,5p,times,twocolumn,number]{elsarticle}

\usepackage{amssymb}
\usepackage{lipsum}
\usepackage{textgreek}
\usepackage{mathtools}
\usepackage{graphicx}
\usepackage{bm}
\usepackage{dcolumn}
\usepackage{multirow}
\usepackage{overpic}
\usepackage{caption}
\usepackage{subcaption}
\usepackage{chemformula}
\usepackage{comment}
\usepackage[most]{tcolorbox}
\usepackage{xcolor}

\begin{document}

\begin{frontmatter}



\input{affliations}

\begin{abstract}
Long-horizon physics-based simulations of battery degradation provide mechanistic insight but remain computationally expensive, limiting their use for dense exploration of operating conditions over extended cycle life. Here, we propose a hybrid physics-probabilistic learning framework for surrogate modeling of lithium-ion battery degradation trajectories at unseen charging rates. Cycle-resolved degradation data generated with a DFN/P2D electrochemical model in PyBaMM are first transformed into capacity-aligned voltage and derivative features and encoded using a Variational Autoencoder (VAE). The resulting two-dimensional latent space organizes degradation trajectories according to both cycle progression and charging protocol. A sparse multitask Gaussian process (GP) is then trained in this latent space using cycle number and C-rate as input variables, providing continuous interpolation of latent degradation dynamics together with posterior uncertainty estimates. Under protocol-level holdout evaluation, the latent-space GP accurately recovers unseen C-rate trajectories and exhibits uncertainty behavior consistent with the support of the training data. When queried at unseen interior C-rates, the model generates latent trajectories that remain coherently positioned between neighboring simulated protocols. Decoding the GP-predicted latent states through the frozen VAE decoder yields smooth voltage-capacity evolution, while Monte Carlo propagation of the GP latent posterior through an auxiliary latent to State of Health (SOH) predictor provides uncertainty-aware SOH estimates. The proposed BattVAE-GP framework therefore offers a computationally efficient and uncertainty-aware surrogate for long-horizon degradation modeling, providing a structured basis for extending battery health prediction toward richer operating conditions and future simulation-experiment fusion.
\end{abstract}

\end{frontmatter}




\section{Introduction}
Battery degradation remains a central obstacle to electrification and renewable-energy integration, because capacity fade directly limits lifetime, safety margins, and total cost of ownership, while the limited ability to estimate, anticipate, and control degradation under realistic operating conditions continues to hinder reliable battery deployment. Lithium-ion batteries dominate portable electronics and are rapidly expanding across electric mobility and grid storage due to their high gravimetric/volumetric energy density, comparatively long cycle life, and low self-discharge \cite{tarascon2001issues, whittingham2012history}. However, the electrochemical complexity that enables the high performance of lithium-ion batteries also complicates degradation prediction, as capacity loss emerges from multiple interacting mechanisms whose kinetics and relative dominance depend strongly on operating conditions, cell design, and usage history \cite{temprano2024advanced}. As a consequence, forecasting degradation under realistic fast-charging regimes, especially outside the range of observed operating conditions, remains unreliable in many practical settings.

Capacity degradation in Li-ion battery cells is typically discussed through two coupled perspectives: calendar aging and cycle aging. Calendar aging captures time-dependent performance loss during storage, influenced by temperature, state-of-charge, and rest conditions. Cycle aging, in contrast, reflects irreversible changes induced by repeated charge–discharge operation and is often accelerated by high cycling rates (C-rates). Mechanistically, lithium-ion battery degradation arises from coupled parasitic and mechanical processes, including electrolyte reduction and solid-electrolyte interphase growth, electrolyte decomposition, lithium plating and subsequent dead-lithium formation, particle cracking, and loss of active material. At the cell level, these coupled degradation processes are commonly expressed through loss of lithium inventory (LLI), loss of active material (LAM), and impedance growth, ultimately limiting the cyclable lithium inventory and/or the electrochemically accessible electrode capacity \cite{D2CP00417H,birkl2017degradation_diagnostics,weiss2021fast,sah2023insight}. The challenge is compounded by strong cell-to-cell variability and the fact that several mechanisms can be simultaneously active and strongly coupled, making it difficult to attribute observed capacity fade to a single causal factor using experimental measurements alone.

Fast charging illustrates this identifiability problem in a particularly acute form. High C-rates are essential for usability and adoption, but changing the C-rate does not “turn on” only one process; it shifts the balance among transport limitations, reaction kinetics, and parasitic side reactions \cite{weiss2021fast}. In experimental datasets, the isolated effect of C-rate on long-horizon degradation is further obscured by multiple confounding factors, including temperature excursions, rest periods, control strategies, measurement noise, and manufacturing variability. This motivates complementary approaches that can (i) control conditions to isolate specific factors and (ii) generalize beyond discrete tested protocols while providing calibrated uncertainty.

Physics-based simulation offers a principled route to controlled “what-if” analysis. In this work, we use PyBaMM \cite{sulzer2021python} as a high-fidelity battery modeling framework, with the Doyle-Fuller-Newman (DFN) \cite{DFN} model as the core electrochemical description. The DFN model is based on porous-electrode theory and pseudo-two-dimensional (P2D) formulations, resolving coupled ion transport, charge conservation, and interfacial reaction kinetics across the cell thickness and within electrode particles. Here, it is configured for an NMC811-Graphite-based Li-ion chemistry and augmented with relevant degradation submodels to represent the major capacity-fade mechanisms under cycling. By simulating cycling at multiple charging rates, the model enables systematic interrogation of C-rate effects under consistent and repeatable conditions. However, resolving degradation trajectories over thousands of cycles with physics-based models remains computationally expensive, which makes dense sampling over a continuum of C-rates and the resulting exploration of unseen operating regimes costly.

To overcome this bottleneck, we propose a hybrid physics-generative modeling pipeline that couples controlled PyBaMM simulations with probabilistic generative learning. We first generate a dataset of long-horizon cycling trajectories (up to 5000 cycles) across a set of discrete C-rates. We then train a variational autoencoder (VAE) on degradation trajectories generated using the DFN model implemented in PyBaMM, with the aim of learning a compact and structured latent representation of degradation dynamics. This representation is designed so that the organization of the latent space reflects meaningful variability in capacity fade evolution and associated electrochemical state trajectories. Finally, we place a Gaussian process (GP) model over the learned latent space as a function of C-rate and cycle number, enabling uncertainty-aware interpolation to unseen charging conditions within the sampled operating domain. Decoding the GP-predicted latent trajectories through the VAE yields full lifecycle predictions at unobserved C-rates together with principled uncertainty estimates that grow where data support is sparse.

Unlike direct data-driven regression approaches that map operating conditions directly to capacity or SOH, the proposed framework first learns a structured latent representation of full cycle-resolved voltage-derived trajectories and then performs probabilistic interpolation within this latent space. This separation enables the model to preserve physically interpretable trajectory information while quantifying uncertainty associated with interpolation at unseen operating conditions.

This paper makes three contributions. (1) A controlled simulation dataset of NMC811 cycling trajectories across multiple C-rates using PyBaMM, designed to isolate the effect of C-rate on long-horizon degradation. (2) A VAE-based generative model that learns a low-dimensional, interpretable latent structure for degradation trajectories and supports trajectory synthesis. (3) A GP-on-latents framework that predicts full lifecycle behavior at unseen C-rates while quantifying predictive uncertainty, enabling risk-aware use in downstream battery management and design studies. We also discuss limitations inherent to simulation-trained generative models, most importantly, potential mismatch between simulated and real cells and outline validation pathways and model refinement strategies to improve transfer to experimental data.

\section{Methods}

\subsection{Physics-based modeling}\label{sec:pybamm}
Physics-based cycling data were generated with PyBaMM (Python Battery Mathematical Modelling) \cite{sulzer2021python} using Doyle-Fuller-Newman (DFN/P2D) model \cite{doyle1995design,doyle1993modeling}, which resolves electrochemical transport and interfacial kinetics in the through-thickness direction of the current-collector/negative electrode/separator/positive electrode stack while neglecting in-plane inhomogeneities. Model parameters were initialized from the O'Kane \textit{et al.} parameter set \cite{D2CP00417H}. In addition, the negative-electrode open-circuit potential (OCP) was replaced by the analytical fitted function reported by Chen \textit{et al.} \cite{chen_2020}. The original O'Kane \textit{et al.} parameter set uses an interpolant of experimentally measured OCP data for the entry ``Negative electrode OCP [V]'', which can introduce solver instabilities. We therefore substituted this interpolated representation with the Chen \textit{et al.} fitted OCP function to improve numerical robustness.

PyBaMM provides several thermal submodels that can be coupled to DFN/P2D to obtain thermo-electrochemical simulations. However, the parameter set used here corresponds to the LG M50 cell and primarily supports electrochemical degradation modeling. The available literature does not provide a complete and consistently calibrated set of thermal parameters for this parameterization, such as effective thermal conductivities and heat capacities, which are required to activate the thermal submodels in PyBaMM. While entropic (reversible) heating coefficients as functions of stoichiometry have been reported, complementary measurements needed for thermal closure are typically unavailable. Following this limitation, no thermal submodel is enabled in this work and the temperature is assumed constant throughout cycling. Consequently, the present simulations isolate electrochemical and degradation effects under isothermal conditions and do not capture temperature rise, thermal gradients, or thermally accelerated aging that may occur during practical high-rate charging.

To represent long-horizon capacity fade under repeated cycling, DFN was augmented via PyBaMM's options interface with coupled degradation physics. Specifically, SEI growth was enabled using a solvent diffusion-limited formulation with SEI-driven porosity change; lithium plating was enabled as partially reversible with plating-induced porosity change; mechanical degradation was represented using PyBaMM's particle mechanics option (swelling and cracking in one electrode and swelling-only in the other); and loss of active material (LAM) was modeled as stress-driven, enabling coupling between electrochemical operation, particle mechanics, and capacity loss. Full model options, parameter overrides, and numerical settings are reported in Supplementary Section~S1.4.

\begin{figure}[!h]
  \centering
  \begin{overpic}[width=\linewidth]{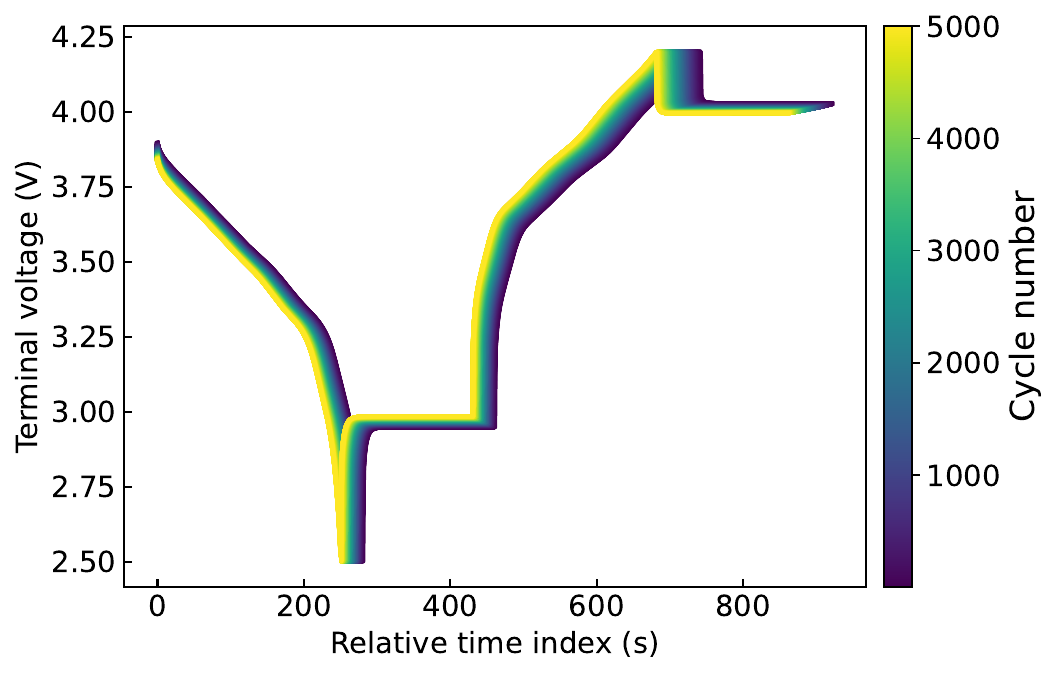}
    \put(30,45){\makebox(0,0){\small Discharge}}
    \put(39,27){\makebox(0,0){\small Rest}}
    \put(60,45){\makebox(0,0){\small Charge}}
    \put(73,56){\makebox(0,0){\small Rest}}
  \end{overpic}
  \caption{Terminal voltage $V$ as a function of relative time $t$ for the simulated cycling protocol at 0.50C. Each curve corresponds to one cycle and is colored by cycle number, illustrating the evolution of the full discharge--rest--charge--rest voltage profile across aging.}
  \label{fig:voltage_vs_time}
\end{figure}
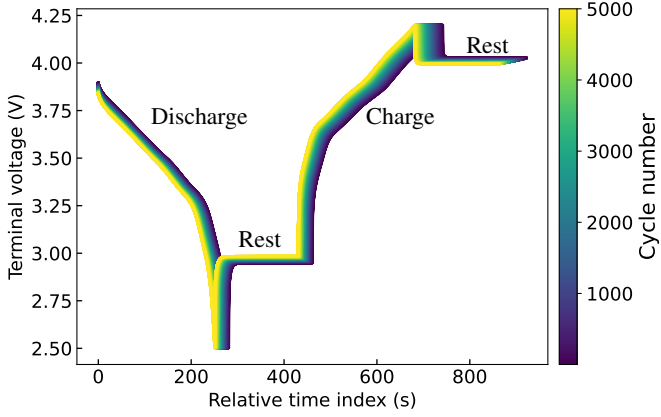

Each simulation followed a repeated discharge-rest-charge-rest sequence with fixed discharge at 1C and a varied constant-current (CC) charge rate $r$, bounded between 2.5~V and 4.2~V, with 30~min open-circuit rests between steps (see Figure \ref{fig:voltage_vs_time}). We simulated $r \in \{0.2,0.3,0.5,0.6,0.75,0.85,1.0\}\,C$ for 5000 cycles, and keep the protocol structure identical across charge rates. Full step definitions and cutoffs are provided in Supplementary Section~S1.2.

As aging progresses, the accessible capacity decreases and the charge passed per cycle changes accordingly. To compare degradation trajectories across C-rates on a consistent basis, degradation indicators were tracked as a function of total charge throughput
\begin{equation}
Q_{\mathrm{tot}}(t)=\int_0^t |I(\zeta)|\,d\zeta,
\label{eq:qtot}
\end{equation}
where \(|I(t)|\) is the magnitude of the applied current.
\subsection{Dataset construction and preprocessing} \label{sec:dataset}

PyBaMM simulations were logged at a fixed sampling period (10~s), exporting per-timestep $(t,V,I,\mathrm{Cycle})$, where
$t$ is time, $V$ is terminal voltage, and $I$ is applied current. Because time-domain sampling produces rate-dependent
sequence lengths (higher C-rate $\Rightarrow$ shorter CC segments), we represent each cycle on a capacity-parametrized
grid and construct fixed-length multi-channel inputs \cite{zheng2018_energy, ZHANG2025124404}.

For each cycle, the charge and discharge CC segments are extracted and mapped to a normalized capacity coordinate
$q\in[0,1]$ using a dataset-level reference capacity $q_0$ (maximum effective CC capacity observed in the dataset). We
define a uniform grid
\begin{equation}
q_n=\frac{n}{N-1},\quad n=0,\ldots,N-1,\qquad Q_{\mathrm{grid}}(q)=q\,q_0,
\label{eq:qgrid}
\end{equation}
and interpolate voltage onto $Q_{\mathrm{grid}}$ using a shape-preserving cubic Hermite interpolant (PCHIP) \cite{Fritsch1984PCHIP}, separately
for charge and discharge, yielding $V_{\mathrm{ch}}(q)$ and $V_{\mathrm{dis}}(q)$.

To place charge and discharge on a consistent SOC-like axis, discharge is aligned using
the achieved end-of-charge fraction $q_{\mathrm{top}}$:
\begin{equation}
\begin{aligned}
q_{\mathrm{top}} &= \min\!\left(\frac{\max(Q_{\mathrm{ch}})}{q_0},\,1\right),\\
Q_{\mathrm{ch}}^{\mathrm{soc}} &= Q_{\mathrm{ch}},\quad
Q_{\mathrm{dis}}^{\mathrm{soc}} = q_{\mathrm{top}}q_0 - Q_{\mathrm{dis}}.
\end{aligned}
\label{eq:soc_align}
\end{equation}

Although the capacity-parametrized voltage curves $V_{\mathrm{ch}}(q)$ and $V_{\mathrm{dis}}(q)$ capture the global
electrochemical response, they are typically smooth and dominated by the underlying thermodynamic OCV, so cycle-to-cycle
changes can be visually subtle. We therefore augment the input representation with derivative-based features commonly used
in differential voltage analysis (DVA) and incremental capacity analysis (ICA). Taking derivatives with respect to charge throughput acts as a local sensitivity operator: $dV/dQ$ highlights changes in slope, and $dQ/dV$ emphasizes regions where
$V(Q)$ exhibits plateaus or rapid curvature changes, revealing sharp and structured features that are often muted in $V(Q)$
itself \cite{dubarry2006ica_ocv, olson2023differential_analysis}. In Li-ion cells, these derivative features are informative because they are closely tied to electrode phase
transitions and the relative alignment of the positive/negative lithiation windows. As degradation progresses, mechanisms
such as loss of lithium inventory (LLI), loss of active material (LAM), and impedance growth typically manifest as
systematic, interpretable perturbations of the derivative signatures (e.g., peak/inflection shifts, amplitude changes, and feature broadening) \cite{dubarry2022ica_best_practices, birkl2017degradation_diagnostics}. Importantly, these signatures are not uniquely attributable to a single mechanism without additional constraints or diagnostics. Here, they are used primarily to provide a more discriminative and sensitive input space for
learning degradation trajectories \cite{dubarry2022ica_best_practices, fly2020rate_dependency_ica}.

Because numerical differentiation amplifies interpolation noise and boundary artifacts, we compute $dV/dQ$
analytically from the shape-preserving PCHIP interpolants. Incremental-capacity features are computed as
\begin{equation}
\frac{dQ}{dV}(q)=\frac{1}{\max(|dV/dQ(q)|,\epsilon)},
\label{eq:ica}
\end{equation}
with a small $\epsilon$ floor to prevent numerical instabilities near plateaus, and mask boundary regions where derivatives are unreliable. All channels are assembled into a fixed-length tensor $x\in\mathbb{R}^{C\times N}$ with a validity mask to exclude undefined regions from the reconstruction loss. Complete preprocessing details (capacity integration, trimming rules, normalization, and representative channel evolution) are provided in Supplementary Section~S1, Supplementary Fig.~S1, and Supplementary Table~S1.


\subsection{VAE model for degradation trajectories}\label{sec:vae}

\begin{figure*}[t]
    \centering
    \includegraphics[width=0.8\linewidth]{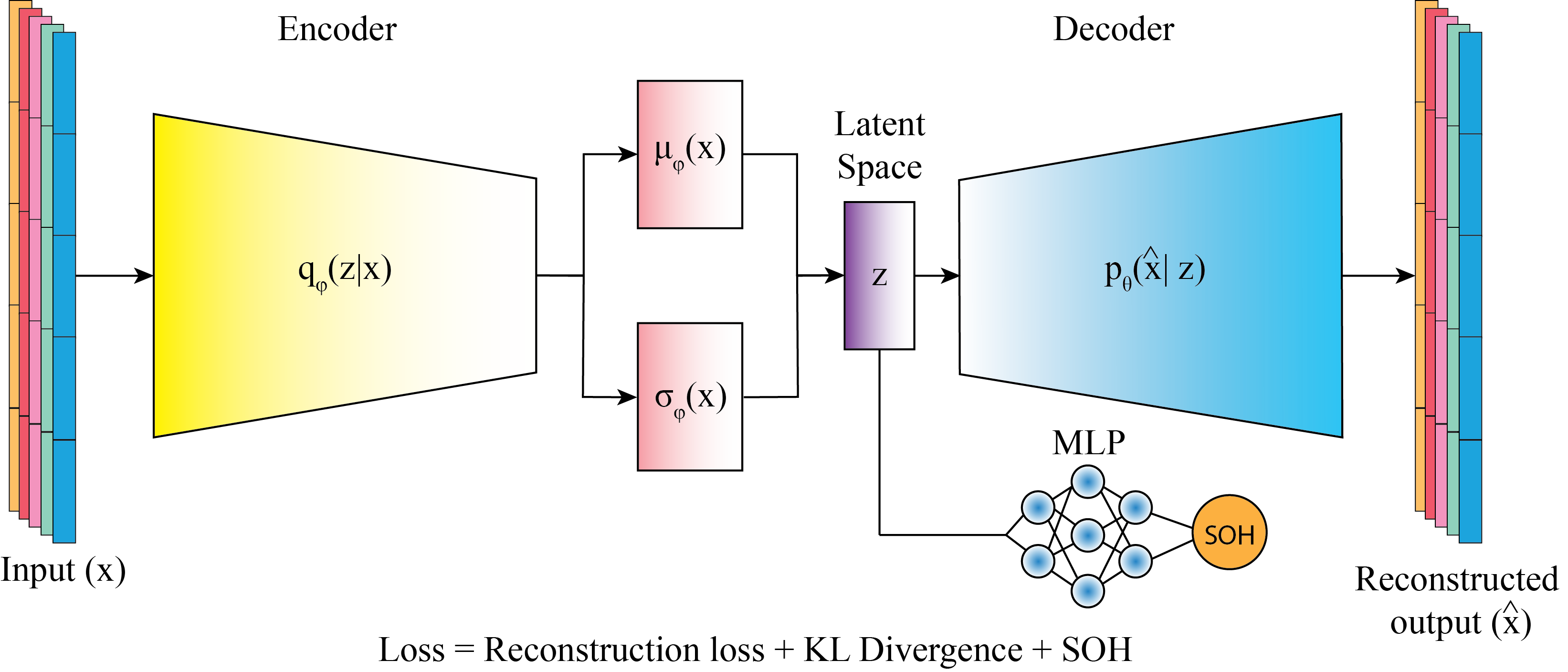}
    \caption{Overview of the proposed Variational AutoEncoder (VAE) framework for battery degradation trajectories. Each simulated cycle, represented after preprocessing as a fixed-length capacity-aligned multi-channel trajectory \(x\), is mapped by the encoder to the parameters of an approximate posterior \(q_\varphi(z\mid x)=\mathcal{N}(\mu_\varphi(x),\mathrm{diag}(\sigma_\varphi^2(x)))\). A latent sample \(z\) is obtained using the reparameterization trick and is used in two parallel branches: a decoder reconstructs the input degradation trajectory, and an auxiliary predictor estimates the corresponding state of health (SOH). The latent representations learned for all cycles are subsequently modeled as a function of charging C-rate and cycle number using a Gaussian process.}
    \label{fig:vae_overview}
\end{figure*}

Following the preprocessing step described above, each simulated charge--discharge cycle is represented as a fixed-length, capacity-aligned multi-channel trajectory
\[
x \in \mathbb{R}^{C\times N},
\]
defined on a normalized capacity grid. In the present work, the input channels comprise the charge and discharge voltage curves and their derivative-based counterparts, namely
\[
\{V_{\mathrm{ch}},\,V_{\mathrm{dis}},\,dV_{\mathrm{ch}}/dQ,\,dV_{\mathrm{dis}}/dQ,\,dQ/dV_{\mathrm{ch}},\,dQ/dV_{\mathrm{dis}}\}.
\]
These channels are designed to capture complementary aspects of cycle evolution: direct voltage response, differential voltage behavior, and incremental-capacity structure. Each individual cycle from each C-rate trajectory is treated as one input example to the model.

The objective of the variational autoencoder (VAE) is to compress each high-dimensional cycle trajectory into a low-dimensional latent variable \(z\) that preserves the dominant degradation information while imposing a structured probabilistic latent space. Such a representation is desirable for two reasons. First, it forces the model to encode the salient variations of degradation trajectories in a compact form rather than memorizing the full input curves. Second, by regularizing the latent space toward a simple prior distribution, it facilitates smooth interpolation and subsequent regression across operating conditions.

We adopt a VAE formulation \cite{kingma2013auto,diederik2019introduction,higgins2017betavae} with a standard multivariate Gaussian prior over the latent variables,
\[
p(z)=\mathcal{N}(0,I)
\]
where \(I\) denotes the identity covariance. For a given input trajectory \(x\), the encoder produces the parameters of an approximate posterior distribution,
\begin{equation}
q_\varphi(z\mid x)=\mathcal{N}\!\left(\mu_\varphi(x),\,\mathrm{diag}(\sigma_\varphi^2(x))\right),
\end{equation}
where \(\mu_\varphi(x)\) and \(\sigma_\varphi(x)\) are the posterior mean and posterior standard deviation, respectively. In this way, the encoder does not map an input cycle to a single deterministic point, but rather to a distribution in latent space. This probabilistic formulation is important because neighboring cycles may exhibit similar but not identical degradation signatures, and the model should be able to represent such variability in a principled manner.

A latent sample is then drawn using the reparameterization trick,
\begin{equation}
z = \mu_\varphi(x) + \sigma_\varphi(x)\odot \varepsilon,
\qquad \varepsilon \sim \mathcal{N}(0,I),
\end{equation}
where \(\odot\) denotes element-wise multiplication. This formulation is used because directly sampling from \(q_\varphi(z\mid x)\) would break the differentiable computational graph needed for gradient-based optimization. By rewriting the stochastic latent variable \(z\) as a deterministic function of \(\mu_\varphi(x)\), \(\sigma_\varphi(x)\), and an auxiliary random variable \(\varepsilon\), the gradients can be backpropagated through the sampling operation during training. The element-wise multiplication by \(\sigma_\varphi(x)\) scales each latent dimension according to its inferred posterior uncertainty, while the additive mean term \(\mu_\varphi(x)\) centers the sample at the location inferred for the input trajectory.

The sampled latent vector \(z\) is used in two parallel learning pathways. In the first pathway, a decoder reconstructs the full multi-channel input trajectory \(\hat{x}\), including all six channels.

This reconstruction objective encourages the latent representation to retain the physically meaningful information needed to reproduce the electrochemical signatures of each cycle. In the second pathway, the same latent vector \(z\) is passed to a multilayer perceptron (MLP) that predicts the SOH associated with that cycle. The inclusion of this auxiliary SOH-prediction branch encourages the latent variables not only to reconstruct the observed trajectories, but also to remain informative with respect to degradation progression.

The model is therefore trained to learn a latent space that is simultaneously reconstructive, probabilistically regularized, and degradation-aware. Reconstruction of the cycle features preserves detailed electrochemical information, Kullback-Leibler (KL) regularization promotes a smooth and structured latent manifold, and the auxiliary SOH prediction links the latent representation to a directly interpretable health metric. This combination is particularly useful in the present setting, where the ultimate goal is not only to compress individual cycle trajectories, but also to obtain a latent representation whose evolution across cycle number and C-rate can later be modeled in a continuous manner by a Gaussian process.

Accordingly, model parameters are learned by optimizing a variational objective based on the evidence lower bound (ELBO),
\begin{equation}
\log p_\theta(x)\ge
\mathbb{E}_{q_\varphi(z\mid x)}\!\left[\log p_\theta(x\mid z)\right]
-\mathrm{KL}\!\left(q_\varphi(z\mid x)\,\|\,p(z)\right),
\end{equation}
which we implement in a \(\beta\)-VAE form \cite{higgins2017betavae} to control the trade-off between reconstruction fidelity and latent-space regularization. In practice, this variational objective is combined with the supervised SOH prediction term so that the latent variables encode both trajectory-level degradation signatures and health-relevant information. The detailed loss formulation and optimization procedure are provided later in the training subsection.

\paragraph{Encoder}

\begin{figure*}[!t]
  \centering
  \includegraphics[
    width=\textwidth,
    height=0.6\textheight,
    keepaspectratio
  ]{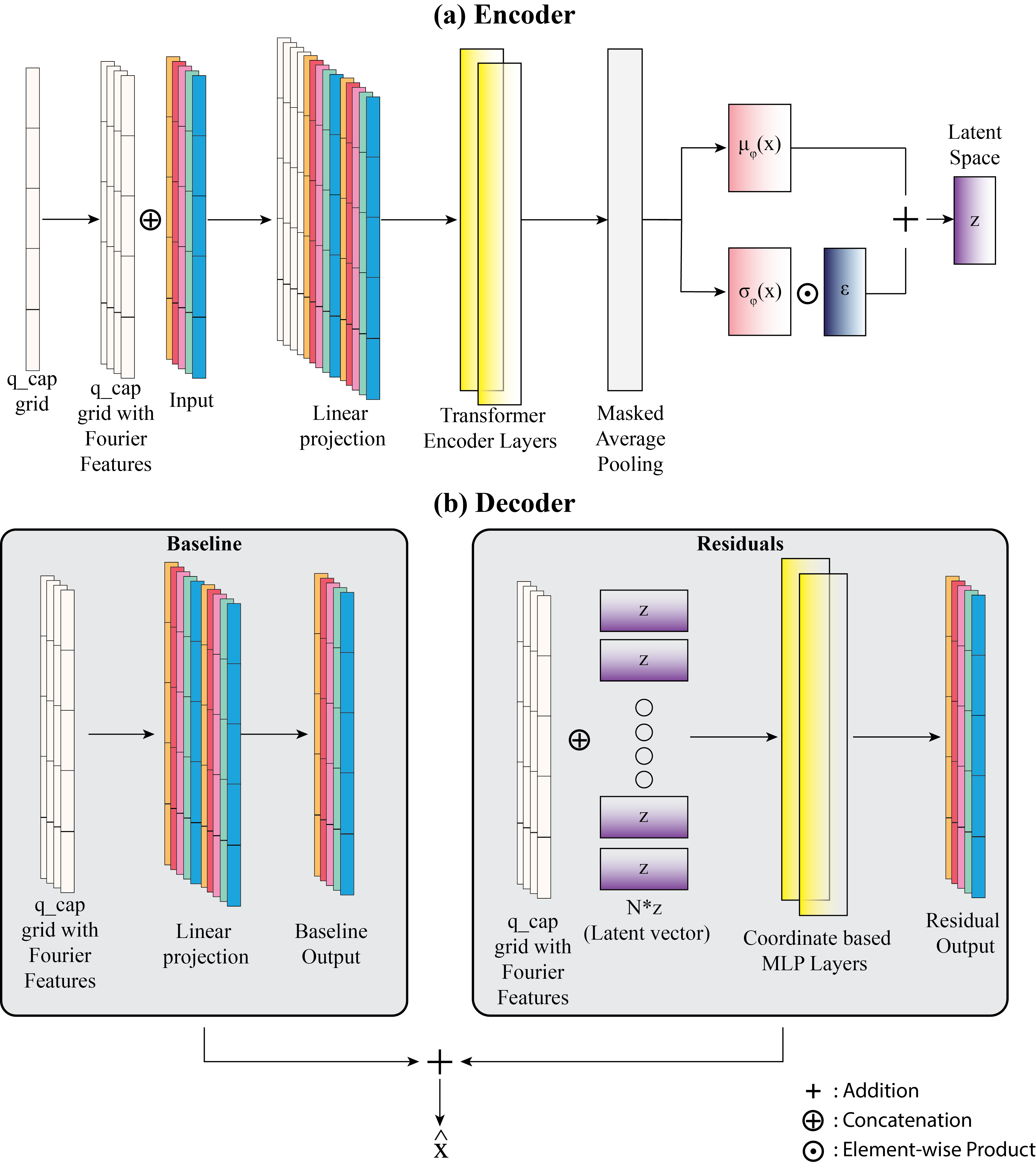}
  \caption{Detailed architecture of the proposed variational autoencoder. (a) Encoder: encoder architecture. Capacity-aligned multi-channel cycle trajectories, together with Fourier features of the normalized capacity grid \(q_{\mathrm{cap}}\), are embedded token-wise and processed by a Transformer encoder with masking to exclude padded or physically invalid grid regions. Masked mean pooling produces a fixed-dimensional trajectory representation, which is then projected to the posterior parameters \(\mu_\varphi(x)\) and \(\log\sigma_\varphi^2(x)\). (b) Decoder: decoder architecture, described in the following paragraph.}
  \label{fig:encoder_decoder_arch}
\end{figure*}

Each degradation trajectory is represented on a fixed \(N\)-point normalized capacity grid \(q_{\mathrm{cap}}\in[0,1]\), so that a cycle is treated as a sequence of \(N\) tokens. For a mini-batch of size \(B\), the encoder input is written as \(x\in\mathbb{R}^{B\times C\times N}\), together with a Boolean validity mask \(m\in\{0,1\}^{B\times N}\), where \(m_{bn}=1\) indicates that the \(n\)-th grid location of the \(b\)-th sample corresponds to a physically valid portion of the trajectory and \(m_{bn}=0\) denotes padded entries. Such invalid entries arise because, under degradation, not all channels remain physically meaningful over the full normalized grid, while the learning architecture nevertheless requires a uniform tensor size across samples. The mask is therefore used to ensure that padded or unphysical regions do not contribute spuriously to feature extraction or sequence aggregation.

In the implemented encoder (see Fig.~\ref{fig:encoder_decoder_arch}a), token features are constructed from the capacity-aligned channels together with a deterministic encoding of the grid coordinate \(q_{\mathrm{cap}}\) itself. To enrich the positional representation, the normalized capacity coordinate is expanded using fixed Fourier features \cite{tancik2020fourier},
\begin{equation}
\begin{aligned}
\varphi(q) = \big[\, q,\; &\sin(2\pi q),\;\cos(2\pi q),\;\ldots, \\
                      &\sin(2\pi n_F q),\;\cos(2\pi n_F q)\,\big]
\in \mathbb{R}^{2n_F+1}.
\end{aligned}
\end{equation}
with \(n_F=4\) in the present implementation, giving a 9-dimensional coordinate embedding. These features provide the encoder with a richer representation of local position along the capacity axis than a raw scalar coordinate alone, which helps the network capture structured variations in the trajectory as a function of normalized capacity. 

Rather than concatenating all physical channels directly at the input, the encoder first applies separate low-dimensional projections to voltage and derivative groups. Specifically, the charge and discharge voltages \([V_{\mathrm{ch}},V_{\mathrm{dis}}]\) are passed through a small MLP, while the derivative-related channels \([dV_{\mathrm{ch}}/dQ,dV_{\mathrm{dis}}/dQ,dQ/dV_{\mathrm{ch}},dQ/dV_{\mathrm{dis}}]\) are processed by a second MLP (Multi-Layer Perceptron). This design reflects the different numerical character of raw voltage and derivative-based features and allows each group to be embedded into a more suitable intermediate representation before fusion. The resulting voltage embedding, derivative embedding, and Fourier coordinate features are then concatenated token-wise and projected linearly to the model dimension \(d_{\mathrm{model}}=64\). 

Before this projection, padded tokens are explicitly zeroed using the validity mask. This masking step improves numerical stability by preventing arbitrary padding values from being interpreted as meaningful inputs. The sequence of token embeddings is then processed by a Transformer encoder \cite{vaswani2017attention} composed of four stacked "TransformerEncoderLayer" blocks with four attention heads and GELU nonlinearities. Self-attention is applied with a key-padding mask derived from the validity mask \(m\), so that invalid grid locations are excluded from attention operations. In the present setting, this masking has a clear physical interpretation: regions that become undefined because of degradation-induced truncation should not participate in the learned contextual interactions across the sequence. The Transformer therefore aggregates only the valid parts of the degradation trajectory while still operating on a fixed-length representation. 

After the final Transformer block, the token representations are passed through a final layer normalization and then reduced to a fixed-dimensional cycle embedding by masked mean pooling. Concretely, if \(h_n\in\mathbb{R}^{d_{\mathrm{model}}}\) denotes the encoded representation of token \(n\), the pooled representation is computed as
\begin{equation}
h_{\mathrm{pool}}
=
\frac{\sum_{n=1}^{N} m_n\, h_n}{\sum_{n=1}^{N} m_n + \epsilon},
\end{equation}
where \(\epsilon>0\) is a small constant for numerical stability. Thus, the pooled embedding is the average of only the valid token representations, rather than an average over the entire padded sequence. This choice ensures that the global cycle representation reflects only the physically meaningful support of the trajectory. 

Finally, the pooled embedding is mapped through two linear heads to obtain the parameters of the approximate posterior distribution,
\begin{equation}
\mu_\varphi(x)=W_\mu h_{\mathrm{pool}}+b_\mu,\qquad
\log \sigma_\varphi^2(x)=W_\sigma h_{\mathrm{pool}}+b_\sigma,
\end{equation}
where the latent dimension is set to 2 in the reported configuration (see Section S3 in supplementary information). The encoder also returns the Fourier coordinate features so that the decoder can reuse the same capacity-grid representation without recomputation.

\paragraph{Decoder}
The decoder branch, illustrated in Fig.~\ref{fig:encoder_decoder_arch}b, reconstructs each capacity-aligned cycle trajectory using a coordinate based baseline-residual decomposition. More broadly, this follows the spirit of coordinate-based implicit neural representations and latent-conditioned continuous decoders \cite{park2019deepsdf}. Given the latent representation \(z\) and the normalized capacity coordinate \(q\), the reconstructed output is written as
\begin{equation}
\hat{x}(q,z)=f_{\mathrm{base}}(q)+f_{\mathrm{res}}(q,z),
\end{equation}
where \(f_{\mathrm{base}}\) is a baseline branch that depends only on the deterministic capacity coordinate, and \(f_{\mathrm{res}}\) is a residual branch that introduces sample-specific corrections conditioned on the latent code.

In the implementation, the decoder receives the coordinate tensor returned by the encoder but uses only its first channel, corresponding to the raw normalized capacity coordinate \(q\). Fourier features are then recomputed internally within the decoder. This design keeps the baseline branch explicitly tied to the capacity grid itself and restricts it to learning the common coordinate-dependent structure shared across degradation trajectories. The baseline branch is therefore intentionally lightweight: it is not intended to capture cycle-specific or C-rate-specific variations, but rather the mean shape of the reconstructed channels as a function of normalized capacity.

The residual branch is responsible for modeling the deviations from this shared baseline. To do so, the coordinate embedding of \(q\) is projected to the decoder hidden dimension, while the latent vector \(z\) is projected separately and broadcast across all grid points. These two projected representations are then combined and passed through a stack of residual feed-forward blocks of the form
\begin{equation}
h \leftarrow h + \mathrm{MLP}\!\left(\mathrm{LayerNorm}(h)\right),
\end{equation}
with GELU activations and dropout. In this way, the decoder reconstructs each trajectory as the sum of a common baseline component and a latent-dependent correction, so that the latent code is used primarily to represent the subtle differences between trajectories rather than the entire trajectory shape from scratch.

This decomposition is particularly suitable for degradation modeling, since battery aging typically produces gradual and structured changes across cycles rather than abrupt changes in the entire voltage profile (see Figure 1). By assigning the common trajectory trend to the baseline branch, the residual branch can focus on the comparatively small but informative deviations associated with degradation progression.

The decoder predicts all reconstructed channels jointly. In the present implementation, the output consists of two voltage channels and four derivative-based channels,

The final prediction is obtained by summing the outputs of the baseline and residual branches. The normalized capacity coordinate itself is not reconstructed, since it is treated as deterministic. Any padded or physically invalid regions are handled through masking in the loss function rather than by forcing the decoder to fit unsupported targets.

\paragraph{Training objective and optimization}
The model is trained by combining three objectives: masked trajectory reconstruction, KL regularization of the latent posterior, and supervised prediction of the SOH. For an input trajectory \(x\), reconstructed output \(\hat{x}\), validity mask \(m\), latent posterior \(q_\varphi(z\mid x)\), and SOH target \(s\), the total training loss is written as
\begin{equation}
\mathcal{L}(x)=\mathcal{L}_{\mathrm{rec}}(x)+\beta(\tau)\,\mathcal{L}_{\mathrm{KL}}(x)+\lambda_{\mathrm{SOH}}\mathcal{L}_{\mathrm{SOH}}(x),
\end{equation}
where \(\beta(\tau)\) is a training epoch-dependent weight controlling the contribution of the KL regularization term during training, and \(\lambda_{\mathrm{SOH}}\) is a fixed scalar coefficient controlling the contribution of the auxiliary SOH-prediction loss. The individual components are defined as
\begin{equation}
\mathcal{L}_{\mathrm{rec}}(x)=
\frac{\left\|m\odot(x-\hat{x})\right\|_2^2}{\sum m},
\end{equation}
\begin{equation}
\mathcal{L}_{\mathrm{KL}}(x)=
\mathrm{KL}\!\left(q_\varphi(z\mid x)\,\|\,p(z)\right),
\end{equation}
and
\begin{equation}
\mathcal{L}_{\mathrm{SOH}}(x)=
\left\|\hat{s}-s\right\|_2^2.
\end{equation}
Here, reconstruction is evaluated only over valid entries of the capacity-aligned trajectory, so that padded or physically undefined regions do not contribute to the optimization.

In the reported experiments, the SOH-loss weight was fixed to \(\lambda_{\mathrm{SOH}}=10^{-3}\). This value was chosen so that the SOH-prediction term remained comparable in scale to the reconstruction loss, thereby allowing both objectives to influence the training dynamics without the auxiliary SOH term overwhelming the trajectory-reconstruction objective. The SOH prediction branch is implemented as a small MLP operating directly on the latent vector \(z\). Specifically, the sampled latent representation is passed through two hidden layers with LayerNorm and ReLU nonlinearities before being mapped to a scalar SOH estimate. This auxiliary supervision encourages the latent space not only to support accurate reconstruction, but also to remain informative with respect to degradation state. In other words, the latent code is then trained to encode trajectory-level variations that are predictive of battery health rather than merely serving as an unconstrained compression variable.

The KL coefficient \(\beta(\tau)\) was scheduled linearly from zero to a target value of \(5\times10^{-4}\) during the early phase of training. This scheduling is important because enforcing KL regularization too strongly from the beginning can degrade reconstruction quality before the decoder has learned a stable baseline representation. In the present model, baseline branch is trained first and then frozen after the initial phase. Delaying the full effect of KL regularization therefore allows the decoder to first establish the common trajectory structure before the latent space is more strongly regularized. The maximum KL weight is kept small, consistent with the low-variance nature of the degradation trajectories and the need to avoid posterior collapse. This training strategy matches the intended role of the decoder decomposition: the baseline branch captures the shared coordinate-dependent trend, whereas the residual branch learns the smaller cycle-to-cycle and C-rate dependent deviations that encode degradation progression.

Optimization is performed using Adam optimizer with a batch size of 64. The initial learning rate is set to \(10^{-3}\), and the learning rate is subsequently adapted using a plateau-based scheduler. Early stopping is applied based on validation loss, and the best checkpoint is selected according to the minimum validation loss. The latent dimensionality is set to \(d_z=2\), which was found to provide a compact yet effective representation for the present dataset.

The training and validation protocol is defined at the level of charging-rate trajectories rather than by randomly mixing cycles from whole dataset. Specifically, the trajectory corresponding to 0.6C is reserved as the validation set, while the model is trained on the remaining charging conditions, namely 0.2C, 0.3C, 0.5C, 0.75C, 0.85C, and 1.0C. This choice places the validation condition near the interior of the sampled C-rate range and therefore tests whether the learned latent representation and decoder generalize to an unseen intermediate operating condition. Consistent with good evaluation practice, all normalization statistics for voltage and derivative-based channels are computed using the training trajectories only and then reused unchanged for the validation set, thereby preventing information leakage from validation into preprocessing.

\subsection{Gaussian-process regression in latent space}

The VAE provides a compact representation of each cycle trajectory in a low-dimensional latent space, but the central objective of the present work is not merely to reconstruct observed cycles. Rather, the goal is to learn how degradation evolves jointly with charging condition and cycle progression, so that latent states can be inferred at unobserved combinations of C-rate and cycle number. To this end, regression is performed not in the original observation space of multi-channel voltage-derived trajectories, but in the latent space learned by the VAE.

This design is motivated by both statistical and physical considerations. The observation space is high-dimensional, strongly correlated across the normalized capacity grid, and composed of multiple channels with different scales and smoothness properties. Direct regression in that space would require learning a large multi-output mapping with many degrees of freedom, which is both data-inefficient and difficult to regularize. By contrast, the VAE concentrates the dominant degradation-related variation into a compact latent representation, thereby reducing the downstream interpolation problem to a much lower-dimensional setting. The GP stage therefore models the evolution of latent degradation states rather than attempting to regress complete voltage and derivative trajectories directly.

Let \(x_i\) denote the preprocessed trajectory of one cycle, and let the trained encoder map it to the approximate posterior
\[
q_\varphi(z\mid x_i)=\mathcal{N}\!\big(\mu_\varphi(x_i),\mathrm{diag}(\sigma_\varphi^2(x_i))\big).
\]
For GP training, we use the encoder posterior mean as the latent target,
\[
y_i=\mu_\varphi(x_i)\in\mathbb{R}^{d_z},
\]
where \(d_z=2\) in the reported configuration. Each latent target is paired with a two-dimensional input vector
\[
u_i=[n_i,\;r_i]^\top,
\]
where \(n_i\) is the cycle number and \(r_i\) is the C-rate. The GP therefore learns a mapping from the joint degradation-condition space \((n,r)\) to the latent space learned by the VAE.

Formally, the latent representation is modeled as a vector-valued function of the two inputs,
\begin{equation}
f:\mathbb{R}^2\rightarrow\mathbb{R}^{d_z}, \qquad y_i=f(u_i)+\varepsilon_i,
\end{equation}
where \(\varepsilon_i\) denotes observation noise. In practice, the implementation uses a multitask Gaussian process so that both latent dimensions are modeled jointly within one GP framework. This is preferable to fitting completely separate regressors for each latent coordinate because it preserves a unified probabilistic treatment of the learned latent representation and provides shared training and inference machinery for the full latent vector. To make training feasible for the full latent dataset, we use a sparse variational multitask GP with inducing points, thereby retaining the nonparametric flexibility of Gaussian processes while avoiding the cost of exact GP inference.

The covariance structure of the GP is defined through a kernel on the two-dimensional input space. If \(u=(n,r)\) and \(u'=(n',r')\), then the kernel \(k(u,u')\) quantifies the expected similarity between the latent states at two operating points. In the present framework, the kernel encodes the assumption that degradation trajectories evolve smoothly as functions of both cycle number and charging rate. This is an appropriate inductive bias for battery aging, where neighboring operating conditions are generally expected to produce related latent degradation states rather than abrupt discontinuities.

In the reported implementation, the 2D latent-space GP uses a radial basis function (RBF) kernel (see Section~S4 in supplementary information).The RBF kernel assumes that latent degradation states vary smoothly with respect to both charging C-rate and cycle number, with correlations decaying continuously as two operating points become farther apart in the \((r,n)\) input space. The GP hyperparameters were selected through hyperparameter optimization prior to the reported runs. In the final configuration, the kernel lengthscale was initialized to 2.0, the output variance to 2.0, and the observation-noise variance to \(2.5\times 10^{-3}\). These values provided a practically useful balance between smooth latent interpolation and sufficient flexibility to capture the observed rate- and cycle-dependent evolution. For completeness, the implementation also exposes a rational-quadratic shape parameter \(\alpha\); however, in the reported RBF configuration this parameter is inactive and does not affect the fitted model.

A key reason for choosing a GP rather than an ordinary MLP is that the GP provides a predictive distribution, not merely a point estimate. For any query input \(u_\star=[n_\star,r_\star]^\top\), the GP returns a posterior distribution over the latent state,
\begin{equation}
p\!\left(z_\star \mid u_\star,\mathcal{D}\right)
=
\mathcal{N}\!\left(\mu_\star,\Sigma_\star\right),
\end{equation}
where \(\mathcal{D}=\{(u_i,y_i)\}_{i=1}^N\) denotes the latent training set. The posterior mean \(\mu_\star\) provides the best latent estimate under the model, while the posterior covariance \(\Sigma_\star\) quantifies predictive uncertainty. This uncertainty is particularly important in the present application because interpolation quality depends strongly on how well the queried \((n_\star,r_\star)\) region is supported by training data. A standard MLP can be trained to approximate the latent mapping, but by default it provides only deterministic outputs and no principled posterior variance. In contrast, the GP naturally expresses reduced confidence in regions with weaker training support, which is valuable when interpreting reconstructed trajectories and SOH predictions at unseen operating conditions.

From a computational perspective, the GP is also well suited to this setting because the regression problem is low-dimensional in both its inputs and outputs: only two conditioning variables are modeled, and the target is a compact latent vector rather than the full trajectory. This allows the model to exploit a smooth-function prior directly in the relevant degradation variables, instead of spending capacity on rediscovering the structure already compressed by the VAE. The sparse variational formulation, together with inducing-point approximation and stochastic minibatch optimization, makes GP training substantially more tractable than exact GP regression on the full latent dataset.

The GP training objective is based on negative log-likelihood (NLL). Intuitively, NLL measures how probable the observed latent targets are under the predictive Gaussian distribution returned by the model. A good probabilistic regressor should assign high probability to the observed data, which corresponds to a low NLL. Unlike a purely deterministic error metric such as mean squared error, NLL evaluates both the accuracy of the predictive mean and the appropriateness of the predictive variance. Thus, the model is penalized not only when its mean prediction is inaccurate, but also when its uncertainty is poorly scaled, either through overconfidence or excessive dispersion. In the implementation, training is carried out through a variational evidence lower bound, while validation performance is monitored using NLL computed from the predicted mean and variance on held-out data. This makes NLL a particularly suitable criterion for model selection in the present uncertainty-aware interpolation setting.

Once trained, the GP provides latent predictions for unseen combinations of cycle number and charging rate. These latent predictions are then propagated through the frozen components of the VAE framework. Specifically, the GP posterior mean latent trajectory can be passed through the frozen decoder to reconstruct the corresponding multi-channel degradation signals, while the latent posterior can be propagated through the SOH prediction head to estimate battery health and its associated GP-induced uncertainty. In this way, the GP does not operate as a standalone surrogate in observation space; rather, it serves as a probabilistic interpolator in the learned latent space, while the frozen decoder and SOH head translate latent predictions back into physically interpretable outputs.

This separation of roles is central to the proposed method. The VAE is responsible for learning a compact and structured representation of battery degradation trajectories, together with the mapping between latent space and observation space. The GP is then responsible for learning how these latent states evolve as functions of cycle number and charging condition, while quantifying uncertainty in the interpolated latent dynamics. Together, these components enable prediction of degradation trajectories and SOH at unobserved operating points without directly fitting a high-dimensional regression model to the raw trajectory channels.

\section{Results and Discussion}

We first examine the degradation behavior produced by the physics-based simulations before introducing the learned representation. Figure~\ref{fig:discharge_capacity} shows the evolution of discharge capacity for all simulated C-rates, represented both as a function of elapsed time and as a function of cycle number. These two views are complementary, since they emphasize different notions of aging: calendar aging in Fig.~\ref{fig:discharge_capacity}a and cycle-resolved degradation in Fig.~\ref{fig:discharge_capacity}b.

When discharge capacity is plotted against elapsed time, higher charging rates exhibit a faster apparent decline. This trend is expected because cells operated at higher C-rate complete more cycles within the same duration and therefore accumulate electrochemical throughput more rapidly. From this perspective, higher-rate protocols appear to age faster because more cycling activity is compressed into the same time interval.

\subsection{Charge-rate dependence of simulated capacity fade}\label{sec:cap_trend}
\begin{figure}[!h]
  \centering
  \includegraphics[width=0.49\textwidth]{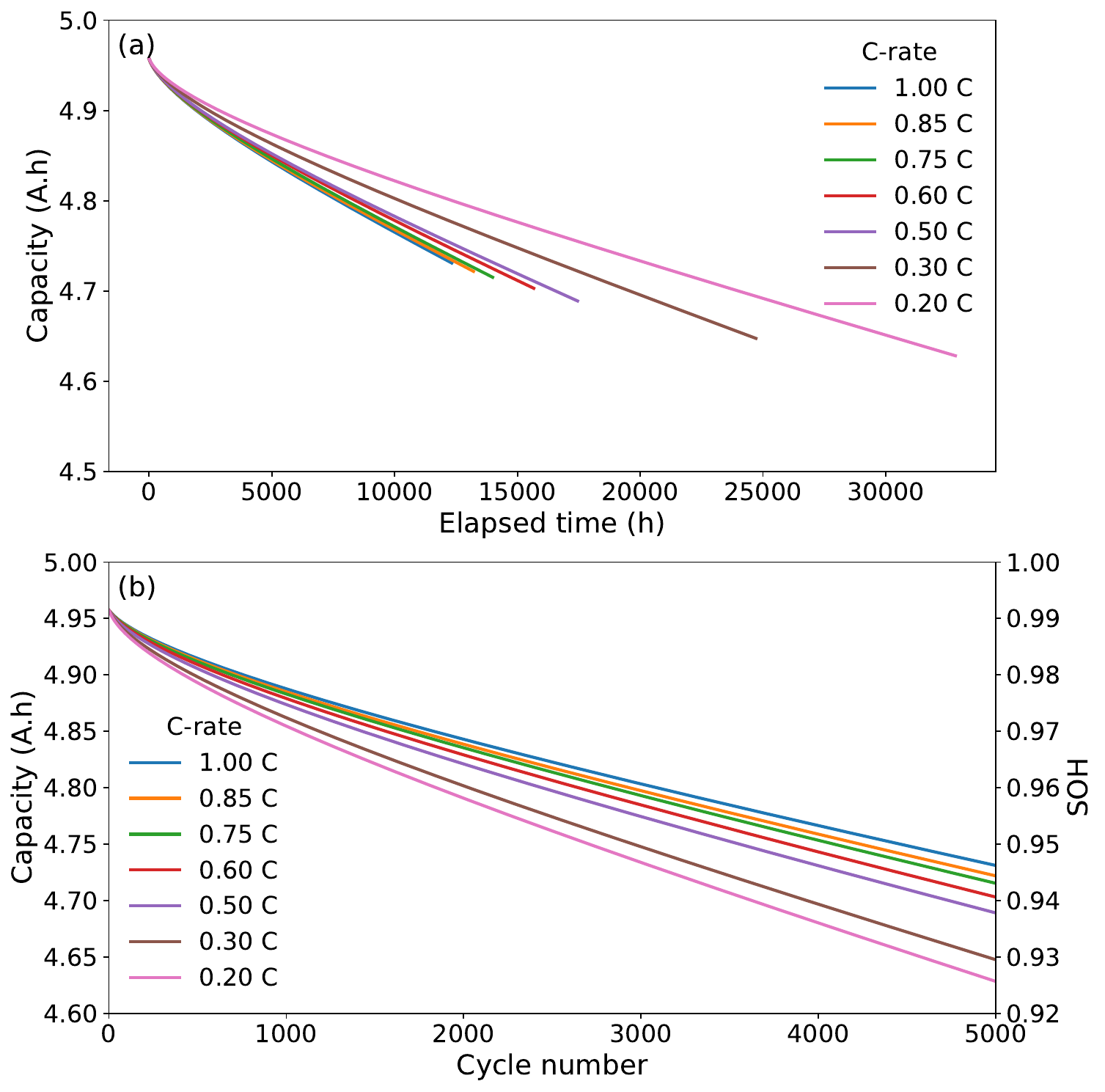}
  \caption{Simulated discharge-capacity evolution across charging C-rates, shown as a function of (a) elapsed time and (b) cycle number. In panel (b), the right-hand secondary axis indicates the corresponding SOH, defined on a cycle-by-cycle basis as $\mathrm{SOH}=Q_{\mathrm{dis}}/Q_{\mathrm{nom}}$, with nominal capacity $Q_{\mathrm{nom}}=5$ Ah.}
  \label{fig:discharge_capacity}
\end{figure}

The cycle-number representation is more directly relevant for the present study, since each cycle is treated as one input sample to the VAE. In this representation, the ordering of the degradation curves changes: lower C-rates exhibit a stronger loss of discharge capacity per cycle over long horizons, whereas higher C-rates retain a larger discharge capacity at the same cycle index. Under the considered protocol, this behavior is physically plausible because lower-rate cycles last longer, so each individual cycle provides more time for side reactions and other degradation processes to accumulate. Thus, although high-rate charging is more aggressive when viewed per unit time, lower-rate charging can lead to higher capacity loss when degradation is compared on a per-cycle basis. This behavior is in line with the previous works \cite{mulpuri2023unraveling, sah2023insight}.

This trend should therefore be interpreted within the assumptions of the present simulation protocol and degradation model. Because temperature is held constant and each lower-rate cycle lasts longer, time-dependent degradation pathways such as SEI growth can contribute more strongly on a per-cycle basis. The observed ordering of the capacity-fade curves therefore reflects the combined influence of the imposed protocol, the selected degradation submodels, and the choice of cycle number as the aging coordinate, rather than a universal statement that lower C-rates are intrinsically more damaging.

A second notable feature of Fig.~\ref{fig:discharge_capacity}b is the non-uniform evolution of capacity over lifetime. The curves display a relatively pronounced early-stage drop followed by a more gradual long-term decline, indicating that degradation does not proceed at a constant rate over cycling. This behavior is consistent with an initial regime of faster degradation followed by a slower and smoother long-horizon evolution. For representation learning, this is important because it indicates that the simulated trajectories are not random or irregular, but instead follow coherent, monotonic degradation paths.

Overall, Fig.~\ref{fig:discharge_capacity} establishes the structure that the representation-learning model must capture. First, the simulated data exhibit systematic and interpretable dependence on C-rate. Second, the degradation trajectories evolve smoothly across cycle life, making them suitable for learning a continuous latent representation. Finally, because the learning problem is posed at the level of individual cycles, the cycle-number view in Fig.~\ref{fig:discharge_capacity}b provides the most relevant physical reference for interpreting the latent-space organization presented next.

\subsection{Latent space organization}

Before analyzing the latent manifold, we verified that the trained VAE accurately reconstructs the multi-channel cycle representations on both training and held-out protocols. The reconstruction errors remained low for voltage channels and captured the main derivative-based features, indicating that the two-dimensional latent representation preserves the dominant degradation signatures needed for downstream GP modeling.

\begin{figure}[!ht]
  \centering
  \includegraphics[width=\linewidth]{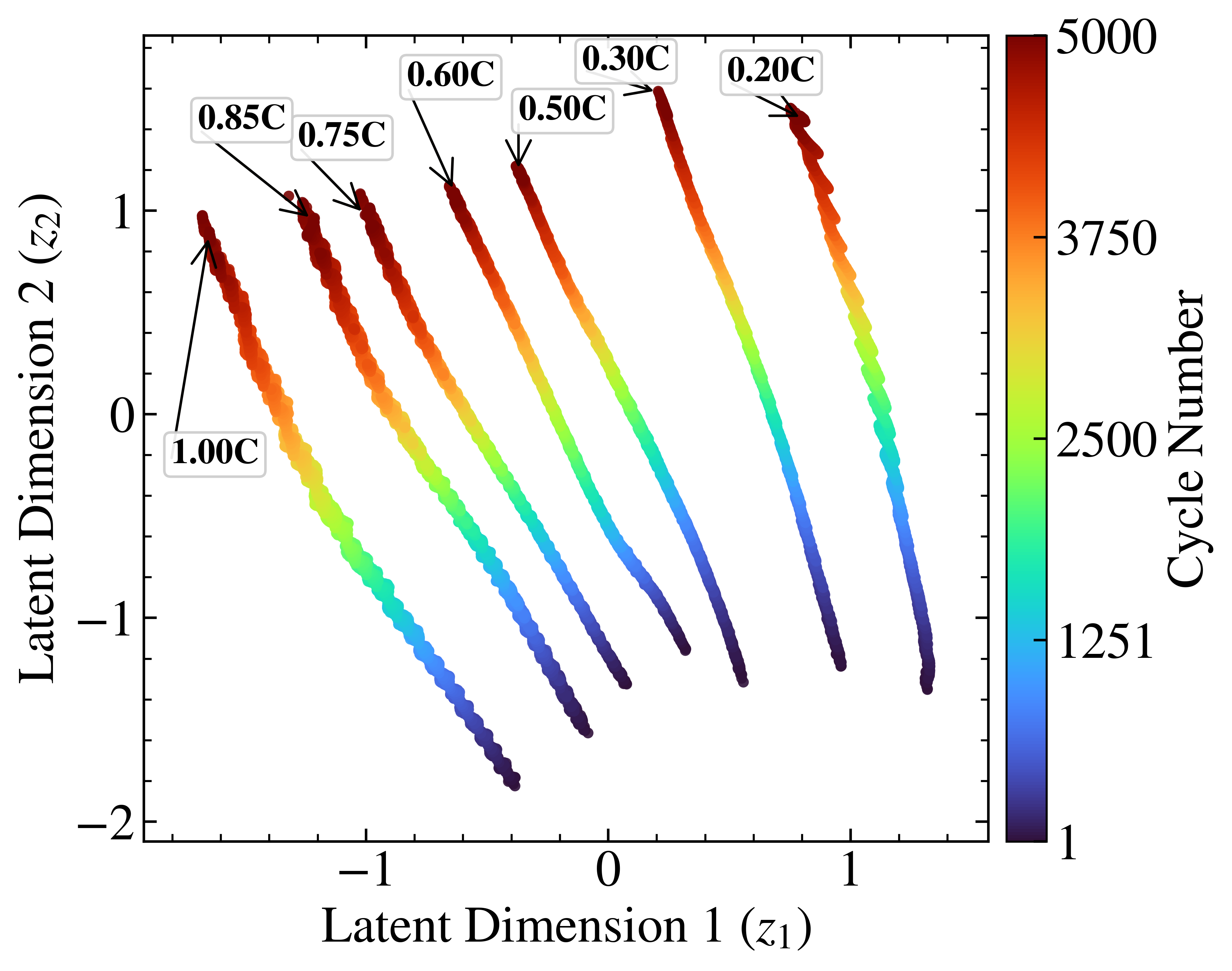}
  \caption{Organization of the learned latent space across charging C-rates. Each point denotes the posterior mean latent embedding, $\mu_{\varphi}(x)$, for a single cycle, with color indicating cycle progression from early to late life. The VAE was trained on the 0.20C, 0.30C, 0.50C, 0.75C, 0.85C, and 1.00C trajectories. The 0.60C trajectory was not used during training and is shown to assess whether the learned manifold supports interpolation across charging conditions.}
  \label{fig:latent_space}
\end{figure}

Having established the cycle-resolved degradation trends in the simulated data, we next examine whether the VAE organizes individual cycles in a latent space that preserves this structure. Figure~\ref{fig:latent_space} shows the posterior mean embeddings, \(\mu_{\varphi}(x)\), for all cycles across the considered charging protocols.

A clear structure emerges in the learned latent space. For each charging protocol, the cycle embeddings evolve along a smooth and narrow trajectory as cycle number increases. This indicates that the encoder does not scatter neighboring cycles arbitrarily, but instead maps them onto a continuous path consistent with gradual degradation progression. Such behavior is desirable because battery aging is expected to evolve through progressive physicochemical change rather than through abrupt, uncorrelated jumps from one cycle to the next.

In addition to the smooth within-protocol evolution, the latent trajectories corresponding to different C-rates are separated in an ordered manner. This separation is consistent with the rate-dependent trends already observed in Fig.~\ref{fig:discharge_capacity}: different charging conditions produce systematically different degradation histories. The learned manifold therefore appears to encode both axes of variation that are central to the problem, namely degradation progression across cycle life and operating-condition dependence across C-rates.

An especially important test of the latent representation is whether an unseen charging condition is mapped to a physically sensible location in latent space. For the held-out 0.60C, the encoded trajectory lies naturally between those of the neighboring observed charging protocols rather than being mapped to an isolated or inconsistent region of the latent manifold. This intermediate placement is a strong indication that the learned latent space supports interpolation across operating conditions. In other words, the encoder appears to organize degradation trajectories continuously with respect to C-rate, rather than merely memorizing the discrete training protocols.

Taken together, these observations suggest that the VAE learns a compact latent manifold that is jointly structured by cycle progression and charging protocol. This property is essential for the subsequent GP stage. A latent space that is smooth within each protocol, ordered across protocols, and well behaved for unseen intermediate conditions is precisely the type of representation in which regression over cycle number and C-rate can be expected to generalize effectively.

\subsection{Gaussian-process modeling of latent degradation trajectories}
\label{sec:gp_results}

Having shown that the VAE organizes degradation trajectories into a smooth and ordered latent manifold, we next examine whether this latent structure can be modeled continuously as a function of operating condition and aging state. To this end, we train a Gaussian-process (GP) model on the latent embeddings produced by the frozen VAE encoder. The goal of this stage is to learn the mapping from charging condition and cycle number to latent state, so that latent trajectories can be inferred at unseen operating points and later decoded back to voltage--capacity trajectories and SOH.

\begin{figure}[!h]
  \centering
  \includegraphics[width=0.45\textwidth]{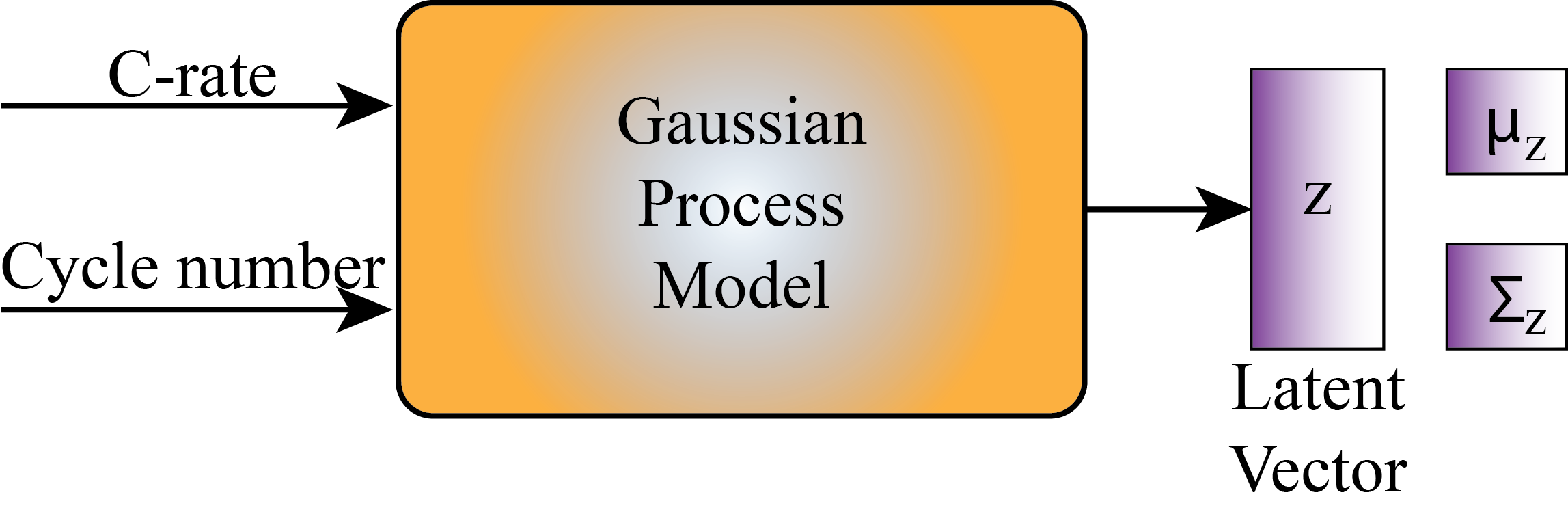}
  \caption{Schematic of the Gaussian-process model in latent space. The two input variables, charging C-rate \(r\) and cycle number \(n\), are used to predict the corresponding latent coordinates \((z_1,z_2)\). The resulting latent predictions are then passed through the frozen decoder and SOH prediction head to reconstruct degradation trajectories and estimate health at unseen operating conditions.}
  \label{fig:gp_model}
\end{figure}

For each cycle trajectory \(x\), the encoder provides a posterior mean latent embedding
\[
\mu_\varphi(x)=(z_1,z_2),
\]
which is used here as the target for the GP model. Each latent point is associated with a two-dimensional input
\[
u=(r,n),
\]
where \(r\) denotes the C-rate and \(n\) denotes the cycle number. The GP therefore learns a two-dimensional regression map from \((r,n)\) to the latent state \((z_1,z_2)\). In the present implementation, this is achieved using a sparse variational multitask GP with inducing points, which is computationally more scalable than an exact GP while retaining a probabilistic predictive distribution. This choice is particularly appropriate in the present setting because the regression inputs are low-dimensional, the latent trajectories evolve smoothly across neighboring rates and cycles, and predictive uncertainty is required for unseen operating conditions. 

 For any query point \(u_\star=(r_\star,n_\star)\), the model yields a posterior distribution over the corresponding latent coordinates. The posterior mean provides the best latent estimate under the model, while the associated uncertainty reflects how strongly that query location is supported by nearby training data in the joint \((r,n)\) space. This is crucial for the present application, since the ultimate objective is not only to interpolate latent trajectories at unseen C-rates.

Because the number of simulated C-rate datasets is limited, the GP must be evaluated in a way that genuinely tests cross-protocol generalization. A random split over cycles would not be meaningful here, since cycles from the same charging trajectory would appear in both training and validation sets, thereby leaking protocol-specific information. We therefore adopt a nested protocol-level holdout strategy. For a chosen outer test C-rate, all cycles from that charging condition are excluded from GP training. Among the remaining non-test C-rates, one rate is further held out as an inner validation fold, and the GP is trained on the rest. This inner leave-one-C-rate-out validation is rotated across all available non-test rates, allowing us to assess the stability of the learned latent-space mapping under repeated protocol-level holdout. After this validation stage, the final GP is retrained on all non-test rates and evaluated on the outer test trajectory. In the present analysis, this procedure is carried out for two outer test conditions, namely 0.60C and 0.75C, in order to show that the conclusions do not depend on a single held-out protocol.

\begin{figure}[!h]
  \centering
  \includegraphics[width=\linewidth, keepaspectratio]{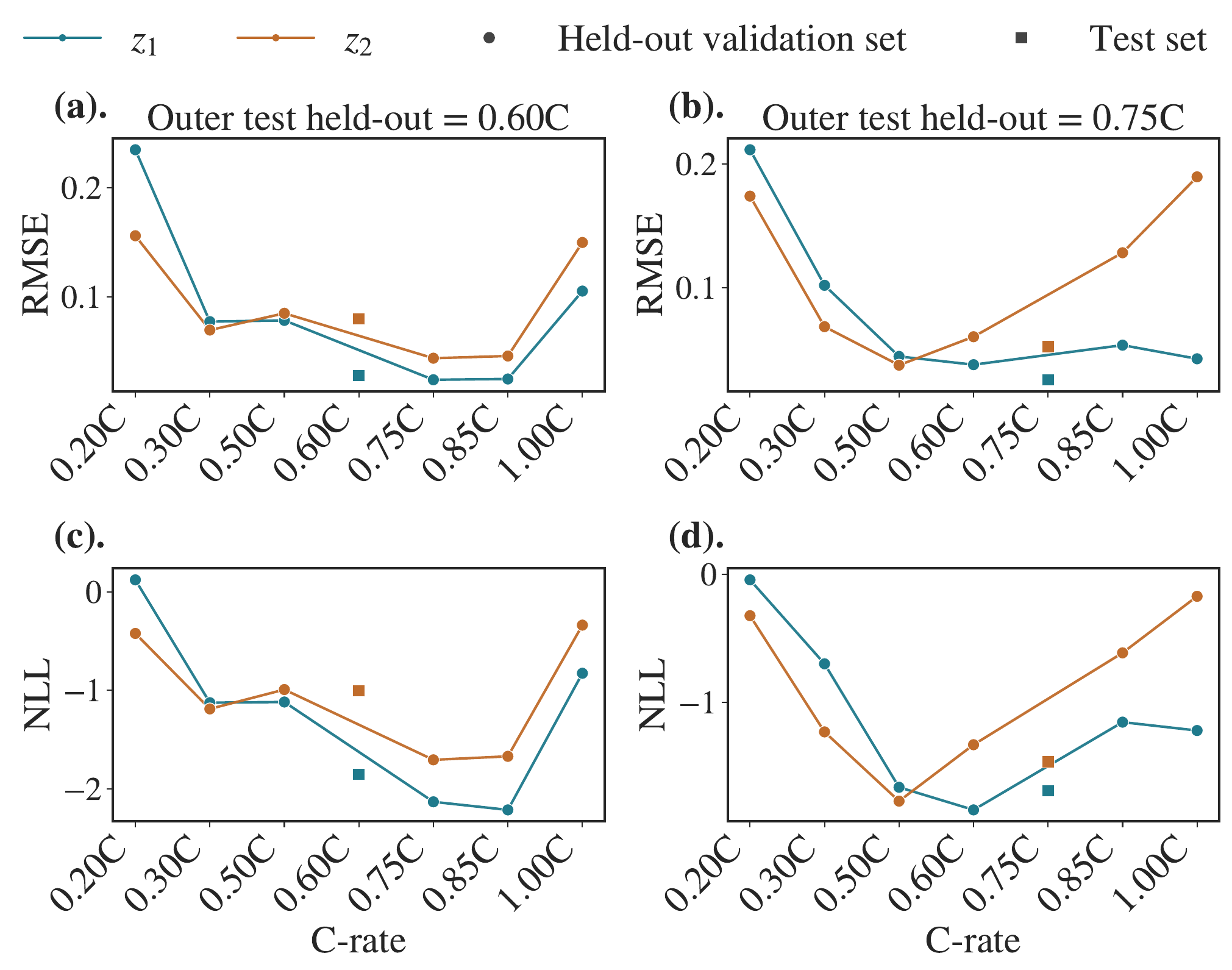}
  \caption{Protocol-level stability analysis of the latent-space GP model under nested C-rate holdout. For each chosen outer test C-rate, one of the remaining C-rates is rotated as an inner validation fold while training on the rest. The figure reports the resulting latent-space prediction error and probabilistic quality for the two latent coordinates \(z_1\) and \(z_2\). After removing the MAE row, the retained panels show RMSE (middle row) and NLL (bottom row) for the inner validation folds, together with the final outer-test result shown by square markers. Panels (a)--(b): RMSE for outer test held out at 0.60C and 0.75C, respectively. Panels (c)--(d): NLL for outer test held out at 0.60C and 0.75C, respectively. Circular markers correspond to inner validation folds, and square markers denote the final evaluation on the outer test trajectory.}
  \label{fig:gp_metrics_by_rate}
\end{figure}

Figure~\ref{fig:gp_metrics_by_rate} summarizes the behavior of the GP model under this nested holdout protocol. The circular markers show the rotated inner validation folds, while the square markers indicate the final outer-test performance for the held-out rates 0.60C and 0.75C. The main purpose of these plots is to assess whether the GP training procedure remains stable when one available C-rate trajectory is removed from training, and then to compare that validation behavior with the final prediction quality on a truly unseen outer test rate.

The retained RMSE panels provide a direct measure of deterministic latent prediction accuracy. They show how closely the GP posterior mean reproduces the VAE-derived latent coordinates for the held-out trajectory. The corresponding NLL panels are particularly important because they evaluate the full probabilistic prediction rather than only the mean trajectory. Negative log-likelihood measures how probable the observed latent targets are under the predictive Gaussian distribution returned by the GP. As such, it penalizes both inaccurate mean predictions and mis-scaled uncertainty. A model that is overconfident and inaccurate is penalized by a NLL close to zero, even if its mean prediction appears visually reasonable, whereas a model whose predictive variance is well calibrated with respect to the observed residuals receives a lower NLL. In the present context, NLL is therefore the more informative metric for deciding whether the GP is suitable as an uncertainty-aware latent interpolator.

To further examine the latent-space predictions, we directly compare the cycle-wise evolution of the GP-predicted latent coordinates with the latent coordinates obtained from the trained VAE encoder, which are used here as reference values. For each outer test condition, the GP is queried across the full cycle range, and the posterior mean and uncertainty band are plotted as functions of cycle number.


Figure~S6 in supplementary information provides a more detailed view of the outer-test behavior for the two latent coordinates. Panels (a) and (b) correspond to the evolution of \(z_1\) with cycle number for the outer test rates 0.60C and 0.75C, respectively, while panels (c) and (d) show the corresponding results for \(z_2\). In all cases, the GP posterior mean follows the VAE-derived reference trajectory closely over the cycle range, indicating that the learned mapping from \((r,n)\) to latent space captures the dominant degradation evolution encoded by the VAE. At the same time, the uncertainty band provides a pointwise estimate of confidence in the inferred latent state.

Taken together, the results in Figs.~\ref{fig:gp_metrics_by_rate} and S6 establish that the latent-space GP is both stable under protocol-level holdout and capable of recovering the cycle-wise evolution of unseen C-rate trajectories with meaningful uncertainty estimates. This provides the necessary confidence to proceed to the next stage of the framework, where the trained GP is queried at intermediate C-rates that were never simulated in the physics-based model, and the resulting latent trajectories are decoded back into voltage-capacity space and SOH predictions.

\subsection{Interpolation of latent trajectories at unseen charging rates}
\label{sec:gp_interp}

Here, we distinguish between heldout C-rates, which correspond to simulated protocols excluded from training and used for quantitative evaluation, and interpolated C-rates, which correspond to operating conditions not included in the original simulation grid and queried only through the trained GP surrogate.

Having established in Section~\ref{sec:gp_results} that the Gaussian-process model can recover held-out latent degradation trajectories under protocol-level validation and test conditions, we next use the trained GP as a continuous surrogate in the joint space of charging C-rate and cycle number. The objective here is no longer validation on an observed charging protocol, but interpolation at operating conditions that were never simulated in the physics-based model.

To this end, we query the trained 2D GP at two unseen charging rates, namely \(r_\star=0.55\)C and \(r_\star=0.70\)C, over the full cycle range \(n=1,\dots,N_{\mathrm{cyc}}\). These two rates were chosen deliberately because they lie in the interior of the simulated C-rate range and are bracketed by neighboring observed protocols. This choice is also consistent with the results of the previous subsection, where latent-space GP performance was found to be most stable in the interior of the available C-rate range, i.e., in the regime where the model benefits from two-sided support from neighboring training trajectories.

For each query pair \((r_\star,n)\), the GP returns a predictive distribution over the latent coordinates \((z_1,z_2)\). In the present subsection, we focus on the posterior mean trajectory in latent space in order to examine the geometric consistency of the interpolated degradation paths. Thus, for each unseen C-rate, the GP generates a continuous latent trajectory
\[
z_\star(n)=\mathbb{E}[z\mid r_\star,n,\mathcal{D}],
\]
which can be compared directly with the latent trajectories obtained from the VAE encoder for the simulated C-rate datasets.

\begin{figure}[!ht]
  \centering
  \includegraphics[width=\linewidth]{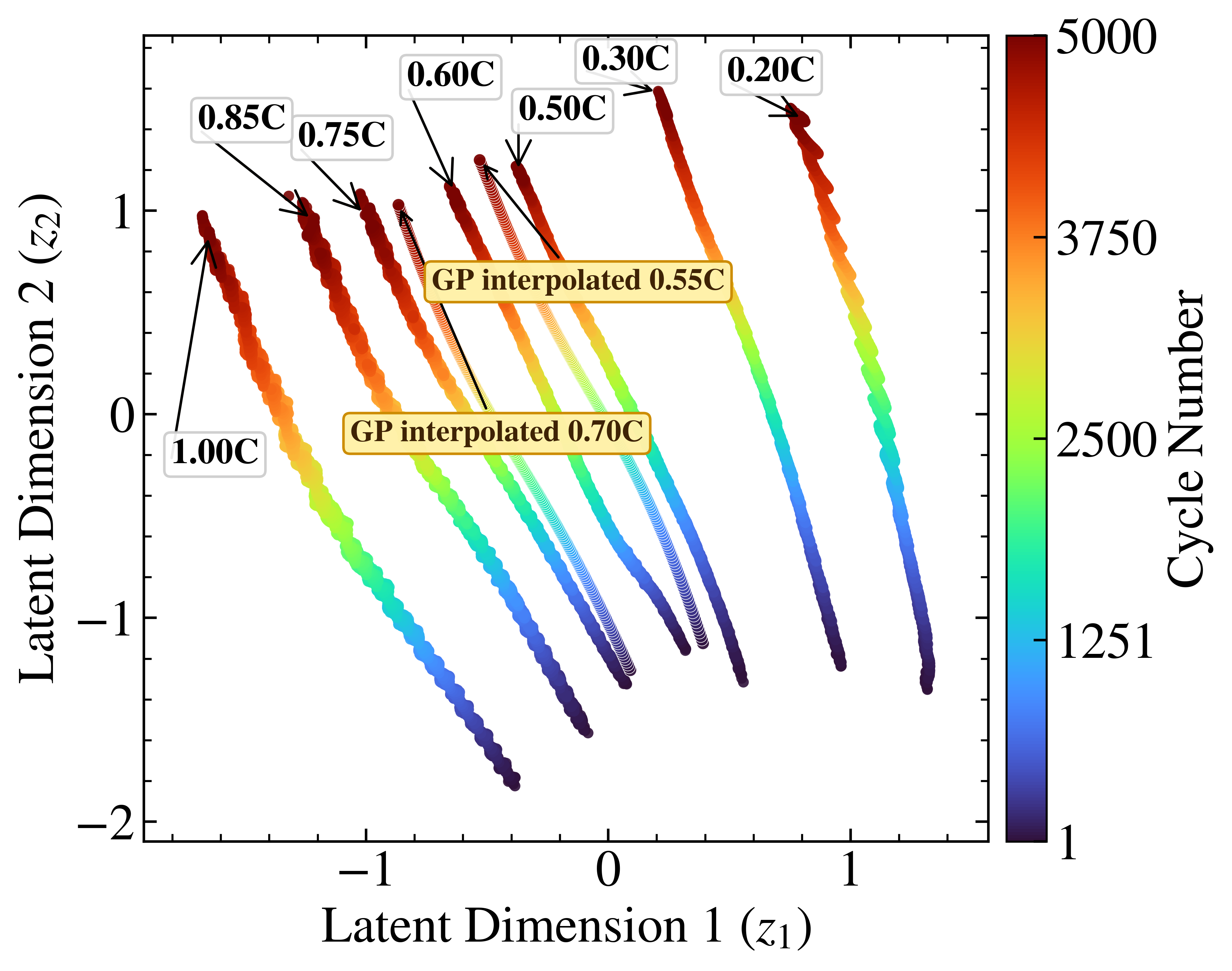}
  \caption{Latent-space organization across C-rates together with GP-interpolated trajectories at unseen interior C-rates. Colored trajectories correspond to VAE posterior-mean embeddings \(\mu_\varphi(x)\) for the simulated C-rates, with color indicating cycle progression from early to late life. The GP-interpolated trajectories at 0.55C and 0.70C are overlaid as additional latent paths.}
  \label{fig:latent_overlay_interp}
\end{figure}

Figure~\ref{fig:latent_overlay_interp} shows the interpolated latent trajectories at 0.55C and 0.70C superimposed on the latent trajectories learned from the simulated C-rate datasets. Several important observations can be made. First, both interpolated trajectories are placed naturally between their neighboring simulated manifolds rather than deviating toward unrelated regions of latent space. In particular, the 0.55C path lies between the 0.50C and 0.60C trajectories, while the 0.70C path lies between the 0.60C and 0.75C trajectories. This is the expected behavior for a physically meaningful interpolation model and indicates that the GP respects the geometry of the latent space learned by the VAE.

Second, the interpolated trajectories preserve the same directional progression with cycle number as the simulated latent paths. The color gradient along each curve shows a smooth evolution from early to late cycles, without discontinuities, reversals, or abrupt shortcuts across the latent manifold. This is important because the latent representation is intended to encode degradation progression, not merely static protocol identity. The fact that the GP-generated trajectories inherit the same progressive structure as the encoded simulated trajectories provides evidence that the GP has learned the underlying degradation dynamics in latent space rather than merely producing geometric averages between neighboring curves.

Third, the interpolated paths are not only intermediate in position but also coherent in shape. Their evolution over cycle life follows the same family of smooth latent deformations observed across the simulated charging protocols. This consistency increases confidence that the latent-space surrogate captures a continuous and physically plausible dependence of degradation behavior on C-rate.

Taken together, these results show that the GP model can be used not only to recover held-out observed trajectories, but also to synthesize latent degradation trajectories at entirely unseen interior charging conditions. Once such latent trajectories have been inferred at unobserved C-rates, they can be propagated through the frozen decoder and SOH predictor to obtain reconstructed voltage-capacity trajectories and health estimates in physically interpretable space.


\subsection{Decoded voltage trajectories and SOH from interpolated latent states}
Having obtained GP-interpolated latent trajectories at the unseen C-rates \(r_\star=0.55\)C and \(r_\star=0.70\)C in Section~\ref{sec:gp_interp}, the final step is to map these latent states back to physically interpretable outputs. In the proposed framework, this is carried out in two complementary ways. First, the GP-predicted latent mean trajectory is passed through the frozen decoder to reconstruct the corresponding voltage-derived channels. Second, the GP latent posterior is propagated through the auxiliary SOH prediction head to quantify how uncertainty in the interpolated latent trajectory translates into uncertainty in state-of-health estimation. This distinction is important because the decoder and SOH head are used differently in the present implementation: voltage reconstruction is deterministic, whereas SOH uncertainty is estimated by Monte Carlo propagation of the GP latent posterior.

We first examine the reconstructed voltage trajectories. For each unseen C-rate and cycle index \(n\), the GP provides a posterior mean latent vector \(\boldsymbol{\mu}_z(n,r_\star)\). This mean latent state is passed through the frozen decoder to generate the corresponding charge and discharge voltage profiles on the normalized capacity grid. In the present pipeline, only the posterior mean latent vector is used for voltage reconstruction; no latent sampling is performed for the decoder, and the reconstructed voltage trajectories therefore remain deterministic. This design keeps the voltage-reconstruction step computationally simple while still allowing the quality of the interpolated latent trajectories to be assessed directly in observation space.


\begin{figure*}[t]
    \centering
    \includegraphics[width=0.9\textwidth]{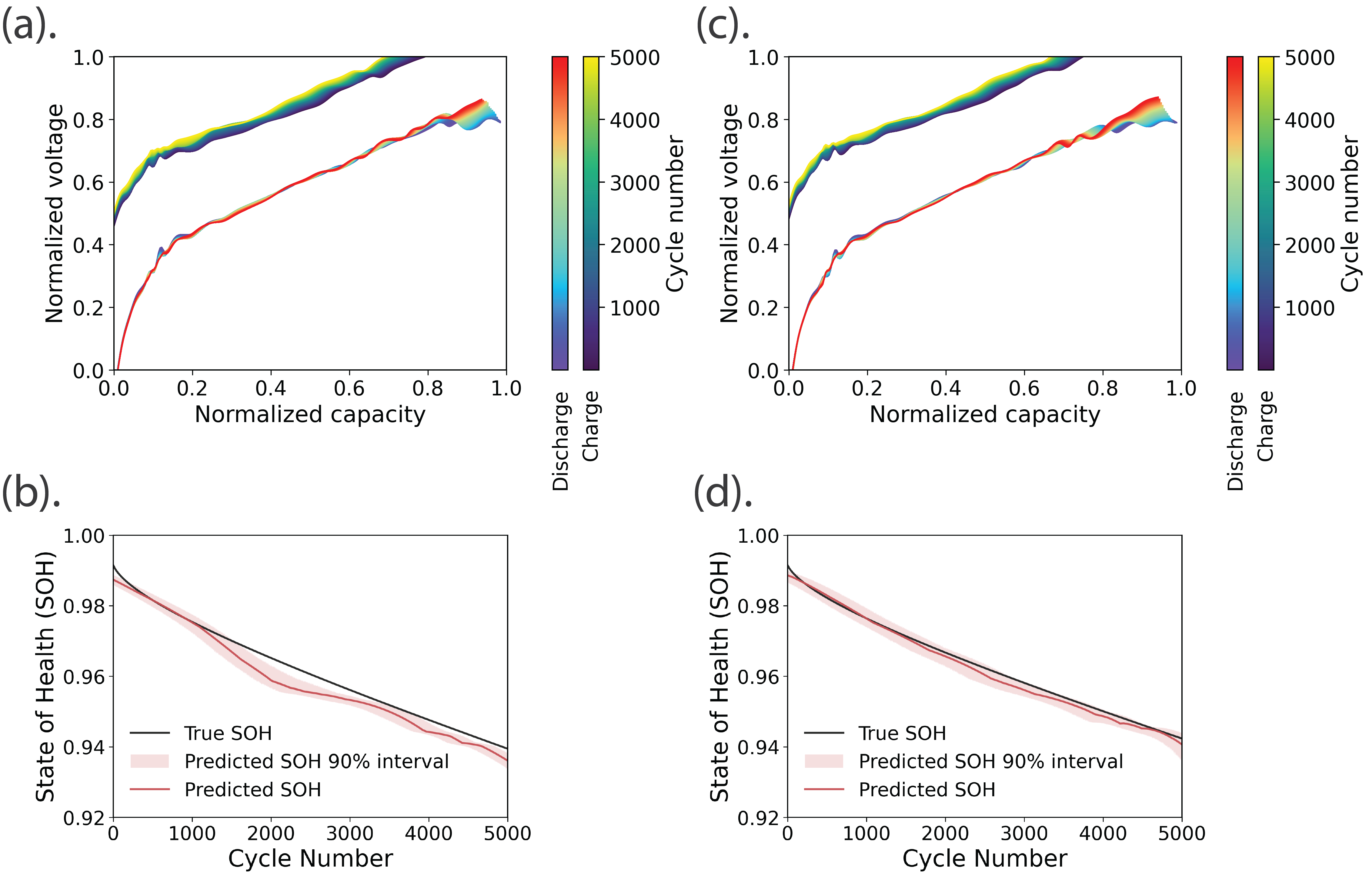}
\caption{ Predictions for two unseen C-rates, \(r_\star=0.55\mathrm{C}\) and \(r_\star=0.70\mathrm{C}\), obtained from GP-interpolated latent trajectories. Panels (a) and (c) show the decoded normalized voltage-capacity trajectories for the two target C-rates. Colors indicate cycle progression from early to late life. Panels (b) and (d) show the corresponding SOH trajectories predicted from the GP-interpolated latent states. The black curve denotes the physics-based reference SOH trajectory, the red curve denotes the deterministic SOH prediction, and the shaded region indicates the empirical 90\% interval window.}
    \label{fig:vae_interpolation}
\end{figure*}

Figure~\ref{fig:vae_interpolation}(a,c) shows that the decoded voltage trajectories remain smooth and coherent over cycle life for both unseen C-rates. The charge and discharge branches evolve continuously from early to late cycles, without abrupt distortions or discontinuities, indicating that the interpolated latent trajectories remain within a region of latent space that the decoder can map back to physically plausible cycle curves. When colored by cycle number, the trajectory families exhibit a gradual progression over lifetime, consistent with the long-horizon degradation behavior observed in the simulated data. Although the C-rates 0.55C and 0.70C were simulated independently, they were excluded from training and used only to assess the quality of the GP-generated unseen-rate predictions. The fact that the decoded voltage profiles preserve the same broad degradation structure as the observed training protocols provides direct evidence that the latent interpolation remains meaningful after being mapped back to the original voltage--capacity domain.

We next assess whether the same interpolated latent trajectories also preserve the information required for SOH prediction. Unlike the voltage-reconstruction step, which uses only the GP posterior mean latent vector, SOH uncertainty is propagated explicitly from the GP posterior in latent space. For each target C-rate \(r_\star\) and cycle index \(n\), the GP provides a two-dimensional Gaussian posterior over the latent state,
\begin{equation}
p(\mathbf{z}\mid r_\star,n)
=
\mathcal{N}\!\left(
\boldsymbol{\mu}_z(r_\star,n),
\boldsymbol{\Sigma}_z(r_\star,n)
\right),
\end{equation}
where
\begin{equation}
\boldsymbol{\mu}_z=
\begin{bmatrix}
z_1\\
z_2
\end{bmatrix},
\qquad
\boldsymbol{\Sigma}_z=
\begin{bmatrix}
\sigma_{z_1}^2 & \mathrm{cov}(z_1,z_2)\\
\mathrm{cov}(z_1,z_2) & \sigma_{z_2}^2
\end{bmatrix}.
\end{equation}
Thus, the GP supplies not only posterior mean estimates for the two latent coordinates, but also their marginal uncertainties and cross-covariance. Retaining this covariance term is important because the uncertainty in \(z_1\) and \(z_2\) is generally correlated, and treating the two coordinates as independent would distort the geometry of the latent posterior.

To propagate this latent uncertainty to SOH, we generate Monte Carlo samples from the bivariate Gaussian posterior at each cycle. This is done by computing the Cholesky factor of the covariance matrix,
\begin{equation}
\boldsymbol{\Sigma}_z = \mathbf{L}\mathbf{L}^{\top},
\end{equation}
where \(\mathbf{L}\) is a lower-triangular matrix obtained from the Cholesky decomposition of \(\boldsymbol{\Sigma}_z\). Let
\begin{equation}
\boldsymbol{\epsilon}^{(s)} \sim \mathcal{N}(\mathbf{0},\mathbf{I}),
\qquad s=1,\dots,S,
\end{equation}
denote independent standard Gaussian samples in two dimensions. A correlated latent sample is then constructed as
\begin{equation}
\mathbf{z}^{(s)}
=
\boldsymbol{\mu}_z
+
\boldsymbol{\epsilon}^{(s)}\mathbf{L}^{\top}.
\label{eq:latent_mc_sampling}
\end{equation}
This transformation maps independent standard-normal noise into correlated latent samples with target mean \(\boldsymbol{\mu}_z\) and covariance \(\boldsymbol{\Sigma}_z\). In other words, the Cholesky factor \(\mathbf{L}\) introduces the correct scaling and correlation structure between \(z_1\) and \(z_2\), so that the sampled latent cloud matches the GP posterior rather than producing independent perturbations along each axis. In the present implementation, \(S=512\) latent samples are drawn for each cycle.

Each sampled latent vector \(\mathbf{z}^{(s)}\) is then passed through the trained SOH prediction head,
\begin{equation}
\widehat{\mathrm{SOH}}^{(s)} = f_{\mathrm{SOH}}\!\left(\mathbf{z}^{(s)}\right),
\end{equation}
where \(f_{\mathrm{SOH}}\) denotes the deterministic MLP learned jointly with the VAE. The main SOH prediction reported in the figures corresponds to evaluating this predictor at the GP posterior mean latent vector,
\begin{equation}
\widehat{\mathrm{SOH}}_{\mathrm{mean}} =
f_{\mathrm{SOH}}\!\left(\boldsymbol{\mu}_z\right),
\end{equation}
while the uncertainty band is obtained empirically from the Monte Carlo sample distribution
\[
\left\{
\widehat{\mathrm{SOH}}^{(1)},\dots,\widehat{\mathrm{SOH}}^{(S)}
\right\}.
\]
In the present work, the shaded band shown in the SOH plots corresponds to the empirical 90\% interval computed from these sampled values. This interval is not obtained by assuming that SOH itself follows a Gaussian distribution, but by nonlinear propagation of the Gaussian latent posterior through the SOH prediction head. Since the SOH MLP is nonlinear, a Gaussian distribution in latent space does not in general remain Gaussian after transformation, and Monte Carlo propagation provides a practical way to capture this effect.

Figure~\ref{fig:vae_interpolation}(b,d) further shows that the SOH trajectories inferred from the GP-interpolated latent states reproduce the expected long-horizon degradation behavior for both unseen C-rates. In both the 0.55C and 0.70C cases, the predicted SOH follows the overall monotonic decline of the corresponding physics-based reference trajectory over the full cycle range, indicating that the interpolated latent states preserve degradation-relevant information not only for reconstructing plausible voltage trajectories, but also for supporting a meaningful scalar description of battery health. This is quantitatively supported by the SOH prediction metrics, with coefficients of determination of (R$^2$=0.9446) for 0.55C and (R$^2$=0.9907) for 0.70C, together with low RMSE values of (3.26$\times10^{-3}$) and (1.26$\times10^{-3}$), respectively. The agreement is particularly encouraging given that SOH is obtained through a second nonlinear mapping from the interpolated latent representation, and therefore constitutes a stricter test of latent-space consistency than visual trajectory reconstruction alone.

A closer inspection of Fig.~\ref{fig:vae_interpolation}(b,d) shows that the predicted SOH curves capture both the global degradation trend and much of the gradual cycle-to-cycle evolution of the reference trajectories. For the 0.55C case, the predicted SOH remains close to the simulation-derived reference over most of the lifetime, with only moderate local deviations and a slight tendency toward underestimation over part of the intermediate cycle range before approaching the reference more closely near end of life. For the 0.70C case, the agreement is similarly strong, with the predicted trajectory closely tracking the reference across most of the cycle horizon and only modest deviations becoming visible in limited regions. In both panels, the empirical 90\% uncertainty interval remains relatively narrow, which is consistent with the fact that these target C-rates lie in the interior of the sampled C-rate range and are therefore supported by neighboring protocols on both sides.

These results are important for two reasons. First, they show that the GP-interpolated latent trajectories remain sufficiently well structured for the auxiliary SOH prediction head to recover a physically meaningful health-evolution signal at unseen charging conditions. Second, they show that uncertainty propagated from the GP latent posterior into SOH space remains of a reasonable magnitude relative to the observed prediction error, rather than becoming unrealistically broad or collapsing to an overconfident point estimate. At the same time, the residual discrepancies remain informative. Since the SOH prediction depends on both the accuracy of the GP interpolation in latent space and the sensitivity of the nonlinear latent-to-SOH mapping, even small local mismatches in the interpolated latent trajectory can produce visible perturbations in the scalar SOH estimate.

The uncertainty band provides an additional layer of interpretation. Because it is obtained by propagating the GP latent posterior through the SOH head, it reflects uncertainty associated specifically with latent interpolation at unseen operating conditions. It should therefore be interpreted as GP-latent-induced SOH uncertainty rather than as total predictive uncertainty of the full framework. In particular, the SOH predictor weights are treated as fixed after training, model-form uncertainty in the VAE and SOH head is not included, and no analogous Monte Carlo uncertainty is propagated through the decoder for the voltage channels. The reported SOH interval therefore quantifies how uncertainty in the interpolated latent trajectory translates into uncertainty in the inferred health evolution, rather than all possible sources of uncertainty in battery aging prediction.

Taken together, the panels in Fig.~\ref{fig:vae_interpolation} provide complementary evidence for the proposed framework. The decoded voltage trajectories show that GP-interpolated latent mean states can generate smooth and degradation consistent voltage-capacity profiles for unseen C-rates, while the SOH results show that the same latent posterior remains sufficiently informative to support uncertainty-aware health prediction. These results complete the end-to-end demonstration of the VAE-GP framework: starting from physics-based degradation simulations, the model learns a compact latent representation, interpolates that representation continuously over C-rate and cycle number, and maps it back to physically interpretable voltage trajectories and health indicators at previously unseen operating conditions.

\section{Conclusion}
\label{sec:conclusion}

In this work, we introduced a hybrid physics--probabilistic learning framework (BattVAE-GP) for modeling long-horizon battery degradation trajectories and generalizing them to unseen charging conditions. Starting from high-fidelity physics-based simulations generated with a DFN/P2D model in PyBaMM, we constructed a learning pipeline that combines capacity-aligned cycle representations, a VAE for latent representation learning, and a Gaussian-process surrogate for continuous interpolation in latent space. The central objective was not only to compress degradation trajectories, but to learn a structured latent representation in which degradation could be modeled continuously as a function of C-rate and cycle number, while retaining a principled description of uncertainty in the interpolated latent dynamics.

The results show that this objective is achievable in a controlled setting. The simulated degradation trajectories exhibited systematic and interpretable dependence on C-rate, and the learned VAE latent space organized these trajectories in a smooth and ordered manner across both cycle progression and charging protocol. This structure proved suitable for downstream probabilistic regression: the latent-space Gaussian process was able to recover held-out C-rate trajectories under protocol-level evaluation, while also providing predictive uncertainty that varied meaningfully with the support of the training data. When queried at unseen interior C-rates, the GP produced latent trajectories that were placed consistently between neighboring simulated rate manifolds, and these interpolated latent states could be decoded into physically coherent voltage-capacity curves and meaningful SOH trends. Taken together, these results support the view that the proposed framework captures a transferable low-dimensional description of degradation dynamics rather than merely memorizing discrete simulated protocols.

An important aspect of the present framework is that uncertainty is handled explicitly, but in a targeted manner consistent with the modeling pipeline. The Gaussian process provides a posterior distribution over the interpolated latent state, which serves as the main source of uncertainty quantification in the framework. In the present implementation, voltage trajectories are reconstructed deterministically from the GP posterior mean latent trajectory, whereas SOH uncertainty is estimated by Monte Carlo propagation of the bivariate latent posterior through the trained SOH prediction head. The reported uncertainty should therefore be interpreted as GP-latent-induced predictive uncertainty, rather than as a complete account of all uncertainty sources in the end-to-end model. Even with this restriction, the uncertainty estimates remain useful because they indicate where interpolation is well supported by neighboring operating conditions and where confidence should be reduced.

In addition, practical implication of the framework is computational efficiency. High-fidelity DFN-based degradation simulations are computationally expensive, particularly when long horizons, multiple operating conditions, and coupled degradation mechanisms are considered. Once trained, the proposed VAE-GP surrogate provides a substantially cheaper way to infer degradation trajectories at unseen C-rates, especially in the interior of the training range where interpolation is best supported. In this sense, the Gaussian-process latent model can serve as a useful surrogate for generating additional protocol-conditioned degradation trajectories, while also supplying a quantitative indication of interpolation confidence in latent space. Such a capability is valuable not only for accelerating data generation, but also for guiding where new expensive simulations or experiments are most needed.

Beyond its immediate predictive utility, this work also provides a structured methodological blueprint for studying battery degradation. Real experimental degradation data are often difficult to interpret because multiple operating factors and degradation mechanisms are coupled simultaneously, obscuring the relationship between observable trajectories and underlying aging dynamics. By first developing and validating the representation-learning and latent-interpolation framework in a controlled physics-based setting, the present study establishes a principled foundation for tackling more realistic and heterogeneous degradation regimes. In this sense, the framework is not only a surrogate model, but also a step toward a more interpretable treatment of complex degradation data.

Several directions naturally follow from this study. First, the operating space can be expanded beyond constant C-rate variation to include more realistic cycling conditions, such as variable current profiles, dynamic charging schedules, temperature dependence, and different depth-of-discharge windows. These factors are known to influence degradation strongly and, when incorporated jointly, would bring the learned latent dynamics closer to real-world battery use. Second, once the model is extended to a richer operating space, simulated and experimental datasets could be incorporated together within a common representation-learning framework. Such a hybrid training strategy would be particularly attractive because physics-based simulations provide controlled coverage of operating conditions, whereas experiments supply the variability and complexity required for practical deployment. Third, with larger and more diverse datasets, it would become meaningful to replace or augment the Gaussian-process latent surrogate with more expressive generative priors, such as diffusion-based latent models~\cite{wehenkel2021diffusion}. These models are attractive because they can represent more complex latent distributions and richer trajectory variability than smooth GP priors, although they also require substantially more data and careful training to remain physically interpretable. A further promising direction is to analyze the geometry of the learned latent trajectories themselves. In particular, quantities such as local curvature, bending direction, and the cycle locations at which these geometric changes occur may provide additional information about transitions in degradation regime. Such geometric markers could help identify whether a trajectory begins to accelerate, deviate, or reorient at later stages of life in a manner consistent with the onset of a new battery degradation mechanism or a change in the dominance of existing mechanisms. In a broader context, this type of latent-trajectory analysis could provide a useful blueprint for studying experimental degradation data, where the underlying mechanisms are often entangled and difficult to isolate directly from raw observables. If robust geometric fingerprints can be associated with specific degradation behaviors, they may help detect, interpret, and ultimately control harmful degradation pathways in practical battery systems. Future work should also focus on mapping the learned latent space onto physically meaningful degradation descriptors. Relating latent coordinates and trajectory geometry to quantities such as loss of lithium inventory, loss of active material, impedance growth, or electrode-utilization changes would strengthen the interpretability of the framework and help connect latent-space evolution to underlying degradation mechanisms.

Overall, the present work demonstrates, in a controlled simulation setting, that long-horizon battery degradation trajectories generated by computationally expensive physics-based models can be distilled into a compact latent representation and interpolated probabilistically across unseen interior operating conditions. The proposed framework provides a computationally efficient route toward uncertainty-aware surrogate modeling of degradation dynamics, while preserving links to physically meaningful outputs such as voltage trajectories and SOH. Future work will extend this approach to broader operating conditions, thermal effects, experimental validation, and simulation–experiment fusion. In addition, replacing or augmenting the GP latent prior with diffusion- or flow-based generative models could enable richer sampling of degradation trajectories, allowing the framework to generate diverse physically plausible aging pathways under complex operating conditions while retaining uncertainty-aware predictions.

\section*{Acknowledgments}
The authors gratefully acknowledge GENCI/IDRIS for providing high-performance computing resources on the Jean Zay supercomputer, which supported the computational aspects of this work. This work was also supported by the CPER MANIFEST and CPER Cornel’IA programs of the Région Hauts-de-France, in the framework of the Predeeption startup project and the TempoBat project.

\section*{Code and data availability}
The code developed for this study is publicly available at "https://github.com/Image-DataScience-Team-LRCS/BattVAE-GP". The data supporting the findings of this work were generated using the PyBaMM battery modeling framework. The processed datasets used for training and evaluation are available from the corresponding author upon reasonable request.

\bibliographystyle{elsarticle-num}
\bibliography{biblio.bib}

\end{document}


\begin{center}
{\Large\bfseries Supplementary Information}
\end{center}
\vspace{1em}

\begin{frontmatter}

\input{affliations}
\end{frontmatter}

\section{Simulation and protocol reproducibility} \label{sec:sim_si}
This section provides sufficient detail to reproduce the PyBaMM simulations used to generate the cycling dataset in the main manuscript.

\subsection{PyBaMM model configuration}
All cycling trajectories were generated with PyBaMM using a porous-electrode Doyle--Fuller--Newman (DFN/P2D) model for an NMC811-based Li-ion cell (LG M50T-type parameterization). Thermal submodels were not enabled and temperature was assumed constant throughout cycling, due to incomplete
availability of thermal parameters required for consistent calibration. To represent long-horizon capacity fade, the DFN model was augmented via PyBaMM's options interface with coupled degradation physics:
\begin{itemize}
  \item \textbf{SEI growth:} solvent-diffusion-limited formulation, with SEI-driven porosity change.
  \item \textbf{Lithium plating:} partially reversible plating, with plating-induced porosity change.
  \item \textbf{Particle mechanics:} swelling and cracking in one electrode and swelling-only in the other.
  \item \textbf{Loss of active material (LAM):} stress-driven LAM, coupled to the mechanics submodel.
\end{itemize}

\subsection{Parameter set and overrides}
Model parameters were initialized from the O'Kane \textit{et al.} parameter set. The negative-electrode open-circuit potential (OCP) was overridden using the corresponding function from Chen \textit{et al.}, by replacing the ``Negative electrode OCP [V]'' entry to match the intended negative-electrode thermodynamics.

\subsection{Cycling protocol and cutoffs}
\begin{figure}[t]
  \centering
  \includegraphics[width=\linewidth]{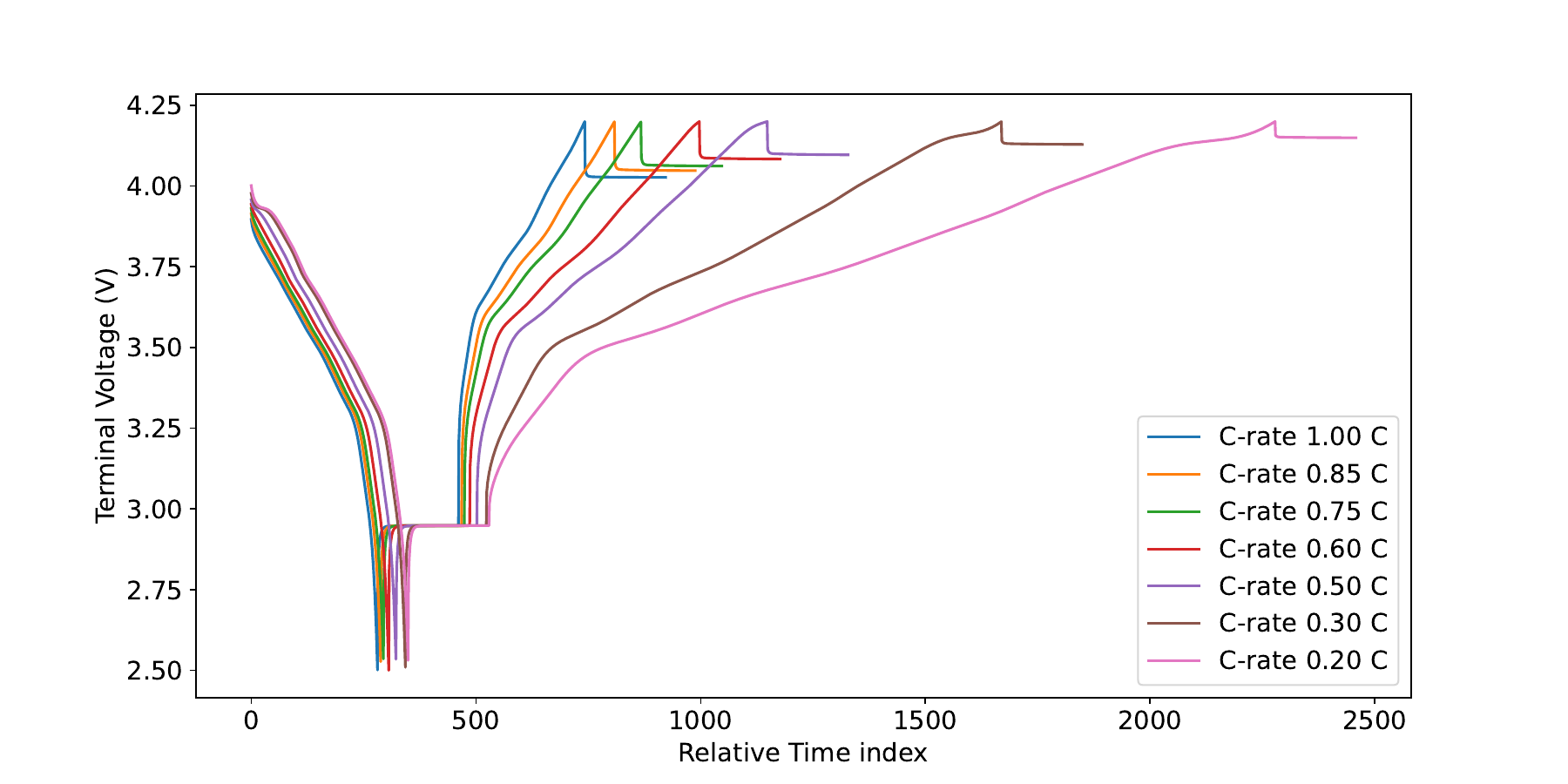}
  \caption{Terminal voltage profiles versus relative time index for representative simulated cycling protocols at different charging C-rates. The discharge–rest–charge–rest trajectories exhibit strong protocol-dependent stretching along the time axis, with lower C-rates producing longer charging segments and higher C-rates yielding more compressed profiles. This mismatch motivates resampling on a normalized capacity grid rather than learning directly from time-indexed voltage signals.}
  \label{fig:raw_data}
\end{figure}
Each simulation followed a repeated discharge--rest--charge--rest sequence designed to isolate the impact of \emph{charge}
C-rate while keeping discharge fixed:
\begin{enumerate}
  \item Initialization at $4.2$ V.
  \item Constant-current (CC) discharge at $1$C to a cutoff voltage of $2.5$ V.
  \item Open-circuit rest for $30$ min.
  \item CC charge at a prescribed C-rate $r$ until $4.2$ V (\emph{no} constant-voltage hold).
  \item Open-circuit rest for $30$ min.
\end{enumerate}
This sequence was repeated for $5000$ cycles. See Figure~S\ref{fig:raw_data}.

Charging rates were simulated for $r \in \{0.2,\,0.3,\,0.4,\,0.5,\,0.6,\,0.75,\,0.8,\,1.0\}\mathrm{C}$, while maintaining discharge at $1$C for all cases. The absence of a CV phase is intentional to keep the protocol consistent across charge rates and attribute differences primarily to CC charging rate.

\begin{table}[t]
\centering
\caption{Cycling protocol used for all simulations (discharge fixed at $1$C; charge rate varied).}
\label{tab:protocol_si}
\begin{tabular}{@{}lll@{}}
\toprule
Step & Condition & Cutoff / duration \\ \midrule
1 & Initialize & $V = 4.2$ V \\
2 & CC discharge & $1$C to $2.5$ V \\
3 & Rest & $30$ min (OCV) \\
4 & CC charge & $r$C to $4.2$ V (no CV) \\
5 & Rest & $30$ min (OCV) \\
\bottomrule
\end{tabular}
\end{table}
\subsection{Numerical settings and solver tolerances}
All simulations were solved using PyBaMM's IDAKLU DAE solver.
Relative and absolute tolerances were set in the simulation configuration (\texttt{rtol} and \texttt{atol}).

Spatial discretization used fixed resolution with:
$$ x_n = x_s = x_p = 10 $$
points through the negative electrode, separator, and positive electrode thickness domains, and
$$ r_n = r_p = 30 $$ points in the negative and positive particle domains.

Simulation outputs were logged at a fixed sampling period of $10$ s, exporting per-timestep records
$(t, V, I, \mathrm{Cycle})$.

\section{Preprocessing and channel construction}\label{sec:preproc_si}

This section provides the complete transformation from raw PyBaMM logs to fixed-length multi-channel cycle
representations used to train the VAE. We report capacity integration, segment extraction, normalized grid construction,
PCHIP interpolation, derivative channel definitions, masking, and normalization.

\subsection{Capacity computation and segment extraction}
PyBaMM outputs are logged on a time grid $\{t_k\}$ with current $\{I_k\}$ and terminal voltage $\{V_k\}$. We compute
cumulative capacity by discrete integration
\begin{equation}
\Delta t_k=t_k-t_{k-1},\qquad
\Delta Q_k=\frac{I_k\Delta t_k}{3600},\qquad
Q_k=\sum_{j\le k}\Delta Q_j,
\label{eq:capacity_discrete_si}
\end{equation}
followed by re-zeroing at the start of each cycle. Charge and discharge CC segments are isolated and re-zeroed per
segment; for discharge, capacity is sign-flipped so that both charge and discharge abscissae are non-negative and
strictly increasing, which is required for stable interpolation.

\subsection{Normalized capacity grid and SOC-like alignment}
A per-dataset reference capacity $q_0$ is estimated as the maximum effective CC capacity $(Q_{\max}-Q_{\min})$ observed
in the dataset. We define a normalized capacity coordinate $q\in[0,1]$ and a uniform grid
\begin{equation}
q_n=\frac{n}{N-1},\quad n=0,\ldots,N-1,\qquad Q_{\mathrm{grid}}(q)=q\,q_0,
\label{eq:qgrid_si}
\end{equation}
where $N$ is the number of interpolation points (reported in Supplementary Table~S1). To compute hysteresis on a common
SOC-like axis, discharge is aligned using the achieved end-of-charge fraction $q_{\mathrm{top}}$:
\begin{equation}
\begin{aligned}
q_{\mathrm{top}} &= \min\!\left(\frac{\max(Q_{\mathrm{ch}})}{q_0},\,1\right),\\
Q_{\mathrm{ch}}^{\mathrm{soc}} &= Q_{\mathrm{ch}},\quad
Q_{\mathrm{dis}}^{\mathrm{soc}} = q_{\mathrm{top}}q_0 - Q_{\mathrm{dis}}.
\end{aligned}
\label{eq:soc_align_si}
\end{equation}

\subsection{Interpolation and channel construction}
Voltage curves are interpolated onto $Q_{\mathrm{grid}}$ using a shape-preserving cubic Hermite interpolant (PCHIP),
separately for charge and discharge, yielding $V_{\mathrm{ch}}(q)$ and $V_{\mathrm{dis}}(q)$.
From these, we construct the multi-channel tensor $x\in\mathbb{R}^{C\times N}$, including:
\begin{itemize}
  \item charge voltage $V_{\mathrm{ch}}(q)$ and discharge voltage $V_{\mathrm{dis}}(q)$,
  \item hysteresis $H(q)=V_{\mathrm{ch}}(q)-V_{\mathrm{dis}}(q)$ on valid overlap,
  \item differential-voltage channels $dV/dQ$ and $d^2V/dQ^2$ obtained analytically from the PCHIP interpolants,
  \item incremental-capacity feature
  \begin{equation}
  \frac{dQ}{dV}(q)=\frac{1}{\max(|dV/dQ(q)|,\epsilon)},
  \label{eq:ica_si}
  \end{equation}
  with $\epsilon>0$ to avoid numerical instabilities in plateau regions where $dV/dQ\approx 0$.
\end{itemize}

\subsection{Masking, trimming, and normalization}
We propagate a validity mask for each cycle/channel to account for incomplete capacity coverage and boundary regions where
derivatives are unreliable. Invalid points are excluded from the reconstruction loss during training. All channels are
normalized using statistics computed on the training split only; the concrete preprocessing hyperparameters (including
$N$, $\epsilon$, trimming widths, and channel-wise normalization) are summarized in Supplementary Table~S1.

\begin{figure}[t]
  \centering
  \includegraphics[width=0.9\linewidth]{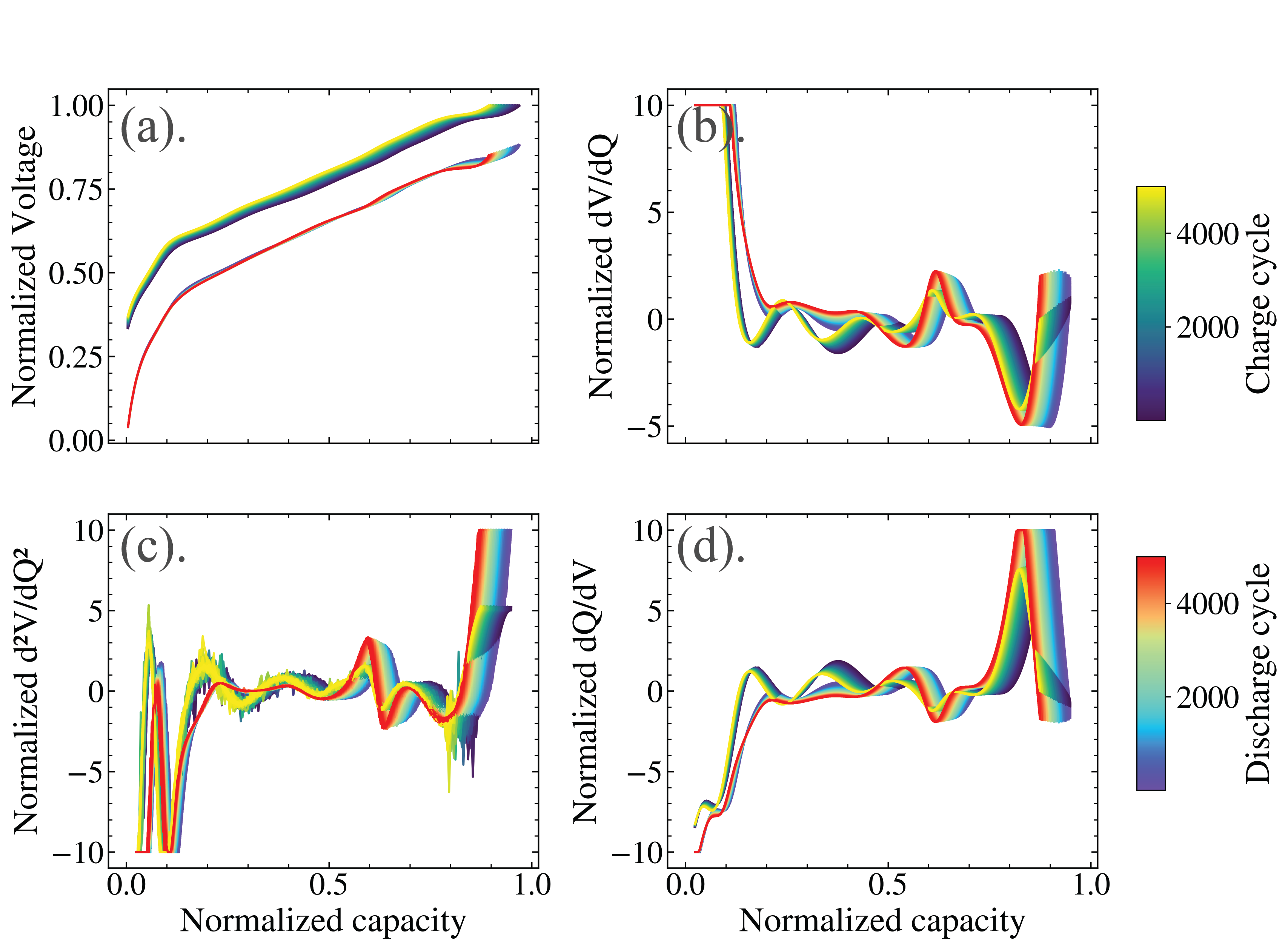}
  \caption{Representative multi-channel inputs constructed from PyBaMM cycles. Voltage and derivative-based channels are
  resampled on a normalized capacity grid and shown for different cycle indices and charge rates.}
  \label{fig:datasets}
\end{figure}

\clearpage
\section{VAE architecture and training details}
\label{sec:si_vae}

This section summarizes the principal VAE hyperparameters and the training workflow used in the main experiments.

\begin{algorithm}[H]
\caption{Training the variational autoencoder on capacity-aligned cycle data}
\label{alg:vae_training}
\begin{algorithmic}[1]
\Require Raw cycle data $\mathcal{D}=\{(t, I, V, n, r)\}$, capacity grid $\mathbf{q}\in[0,1]^N$, encoder $E_\phi$, decoder $D_\theta$, optional SOH predictor $g_\psi$, KL weight schedule $\beta(e)$, number of epochs $E$
\Ensure Trained parameters $\phi^\star,\theta^\star$ (and $\psi^\star$ if SOH head is used)

\State Initialize encoder parameters $\phi$, decoder parameters $\theta$, and optional SOH-head parameters $\psi$
\For{each raw cycle record}
    \State Compute charge/discharge capacity by integrating current over time
    \State Resample voltage curves onto normalized capacity grid $\mathbf{q}$
    \State Compute selected channels, e.g. $V_{\mathrm{ch}}(\mathbf{q}), V_{\mathrm{dis}}(\mathbf{q})$, and optionally derivative channels
    \State Construct binary mask $\mathbf{m}$ for valid entries after interpolation/padding
    \State Store training tuple $(\mathbf{x}, \mathbf{m}, n, r, y_{\mathrm{SOH}})$
\EndFor

\For{epoch $e=1$ to $E$}
    \For{each mini-batch $\{(\mathbf{x}_b,\mathbf{m}_b,n_b,r_b,y_b)\}_{b=1}^B$}
        \State Encode input: $(\boldsymbol{\mu}_b,\log \boldsymbol{\sigma}_b^2) \gets E_\phi(\mathbf{x}_b,\mathbf{m}_b)$
        \State Sample latent variable using reparameterization:
        \[
        \mathbf{z}_b = \boldsymbol{\mu}_b + \boldsymbol{\sigma}_b \odot \boldsymbol{\epsilon}_b,
        \qquad
        \boldsymbol{\epsilon}_b \sim \mathcal{N}(\mathbf{0},\mathbf{I})
        \]
        \State Decode reconstructed channels:
        \[
        \hat{\mathbf{x}}_b \gets D_\theta(\mathbf{z}_b,\mathbf{q},\mathbf{m}_b)
        \]
        
        \State Predict SOH:
        \[
        \hat{y}_b \gets g_\psi(\mathbf{z}_b)
        \]

        \State Compute masked reconstruction loss:
        \[
        \mathcal{L}_{\mathrm{rec}}
        =
        \frac{1}{B}\sum_{b=1}^B
        \mathrm{MaskedLoss}(\hat{\mathbf{x}}_b,\mathbf{x}_b,\mathbf{m}_b)
        \]

        \State Compute KL divergence:
        \[
        \mathcal{L}_{\mathrm{KL}}
        =
        \frac{1}{B}\sum_{b=1}^B
        D_{\mathrm{KL}}\!\left(
        q_\phi(\mathbf{z}\mid \mathbf{x}_b)\,\|\,\mathcal{N}(\mathbf{0},\mathbf{I})
        \right)
        \]

        \State Compute SOH regression loss:
            \[
            \mathcal{L}_{\mathrm{SOH}}
            =
            \frac{1}{B}\sum_{b=1}^B
            \ell(\hat{y}_b,y_b)
            \]
            \State Total loss:
            \[
            \mathcal{L}
            =
            \mathcal{L}_{\mathrm{rec}}
            + \beta(e)\mathcal{L}_{\mathrm{KL}}
            + \lambda_{\mathrm{SOH}}\mathcal{L}_{\mathrm{SOH}}
            \]

        \State Update $\phi,\theta$ (and $\psi$ if applicable) using gradient descent
    \EndFor
\EndFor

\State Save trained encoder $E_{\phi^\star}$, decoder $D_{\theta^\star}$ and SOH head $g_{\psi^\star}$
\end{algorithmic}
\end{algorithm}

\begin{table}[t]
\centering
\caption{Principal VAE architecture and training hyperparameters used in the reported experiments. Entries should be kept consistent with the final configuration used in the main manuscript.}
\label{tab:si_vae_hparams}
\begin{tabular}{@{}ll@{}}
\toprule
Setting & Value \\ \midrule
Capacity-grid points ($N$) & 256 \\
Input channels used by encoder & $V_{\mathrm{ch}},V_{\mathrm{dis}},dV_{\mathrm{ch}}/dQ,dV_{\mathrm{dis}}/dQ,dQ/dV_{\mathrm{ch}},dQ/dV_{\mathrm{dis}}$ \\
Latent dimensionality ($d_z$) & 2 \\
Encoder model dimension & 64 \\
Encoder layers & 4 \\
Encoder heads & 4 \\
Encoder pooling & masked mean pooling \\
Decoder type & additive baseline $+$ residual \\
Batch size & 64 \\
Optimizer & Adam \\
Initial learning rate & $10^{-3}$ \\
Learning-rate schedule & ReduceLROnPlateau \\
KL schedule & linear annealing \\
Baseline freeze epoch & 2 \\
Training termination & early stopping on validation objective \\
\bottomrule
\end{tabular}
\end{table}

\begin{figure*}[!t]
    \centering
    \includegraphics[width=0.88\textwidth]{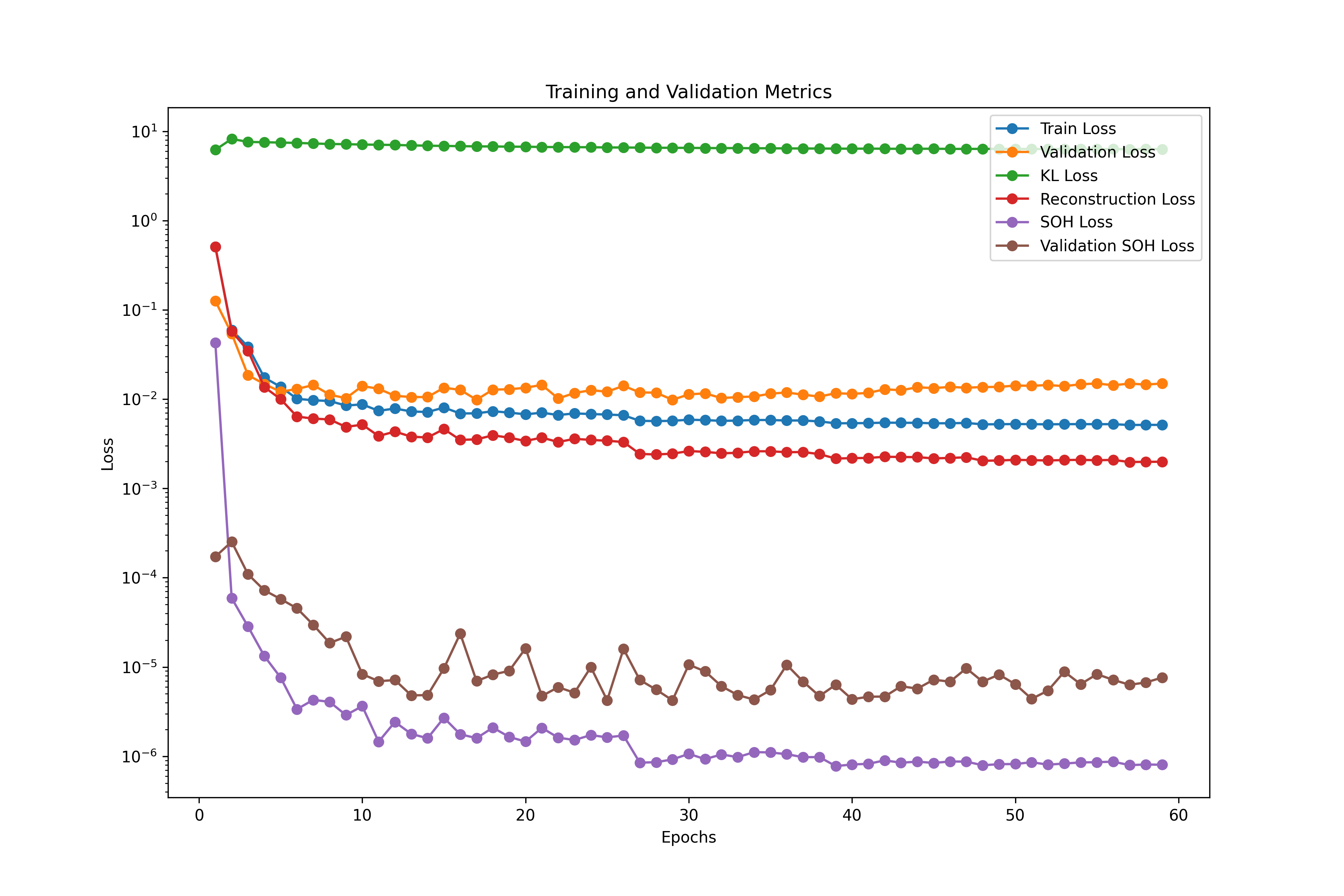}
    \caption{Training and validation metrics over the course of VAE optimization. The total training loss and validation loss decrease rapidly during the first few epochs and then approach a stable plateau, indicating convergence of the optimization. The reconstruction loss shows a consistent downward trend, while the KL term remains comparatively large and varies only mildly, reflecting the regularization imposed on the latent distribution. The SOH prediction loss on both training and validation sets drops by several orders of magnitude and stabilizes at very small values, indicating that the latent representation retains sufficient information for accurate SOH prediction. Overall, the figure shows stable training dynamics, with no evidence of numerical instability and only a modest train--validation gap at convergence.}
    \label{fig:si_training_metrics}
\end{figure*}

\begin{figure*}[!t]
    \centering
    \includegraphics[width=0.6\textwidth]{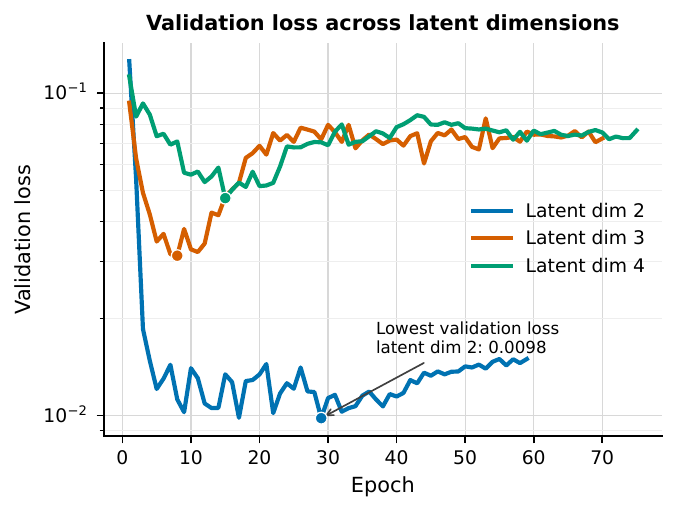}
    \caption{Validation loss comparison for VAE models with latent dimensions 2, 3, and 4. Latent dimension 2 achieves the lowest validation loss, indicating better generalization compared with higher-dimensional latent spaces.
}
    \label{fig:si_validation_loss_latent_dimensions}
\end{figure*}

Since the validation set is constructed from a held-out charging C-rate not seen during training, the validation loss reflects out-of-distribution generalization to an unseen cycling condition. The two-dimensional latent space yields the lowest validation loss, suggesting that it captures the dominant variation associated with C-rate and cycle progression more robustly than higher-dimensional latent spaces, whose additional capacity appears to hurt generalization (see Figure~\ref{fig:si_validation_loss_latent_dimensions}).

\section{Gaussian-process surrogate details}
\label{sec:si_gp}

This section provides additional details on the Gaussian-process (GP) surrogate used to model the latent trajectories learned by the variational autoencoder (VAE). The GP operates in the latent space of the trained VAE and takes charging C-rate and cycle number as the two input variables. For each cycle, the trained encoder provides the posterior mean latent representation, and these posterior mean coordinates are used as regression targets for the GP surrogate.

To preserve the trajectory structure of the degradation process, GP evaluation is performed at the level of complete charging trajectories rather than by random cycle-wise splitting. For a chosen outer-test C-rate, all cycles belonging to that charging condition are excluded from GP training. Among the remaining charging rates, one full trajectory is rotated as an inner validation fold, while the GP is trained on the others. After assessing performance across these protocol-level validation folds, the final GP is retrained on all non-test trajectories and evaluated on the held-out outer-test condition.

\begin{algorithm}[H]
\caption{Fitting Gaussian-process surrogates in latent space}
\label{alg:gp_training}
\begin{algorithmic}[1]
\Require Trained encoder $E_{\phi^\star}$, dataset $\{(\mathbf{x}_i,\mathbf{m}_i,n_i,r_i)\}_{i=1}^M$, latent dimension $d$
\Ensure Trained Gaussian-process models $\{\mathcal{GP}_j\}_{j=1}^d$

\For{$i=1$ to $M$}
    \State Encode the input cycle:
    \[
    (\boldsymbol{\mu}_i,\log \boldsymbol{\sigma}_i^2) \gets E_{\phi^\star}(\mathbf{x}_i,\mathbf{m}_i)
    \]
    \State Define GP input:
    \[
    \mathbf{u}_i = [r_i, n_i]
    \]
    \State Store latent regression target:
    \[
    \mathbf{y}_i = \boldsymbol{\mu}_i
    \]
\EndFor

\For{$j=1$ to $d$}
    \State Construct the scalar training set:
    \[
    \mathcal{T}_j = \{(\mathbf{u}_i, y_{i,j})\}_{i=1}^M
    \]
    \State Initialize GP model $\mathcal{GP}_j$ with the chosen mean and covariance functions
    \State Optimize GP hyperparameters by maximizing the marginal log likelihood
\EndFor

\State \Return $\{\mathcal{GP}_j\}_{j=1}^d$
\end{algorithmic}
\end{algorithm}

\subsection{Hyperparameter optimization of the Gaussian-process surrogate}

\begin{figure*}[!t]
    \centering
    \includegraphics[width=0.95\textwidth]{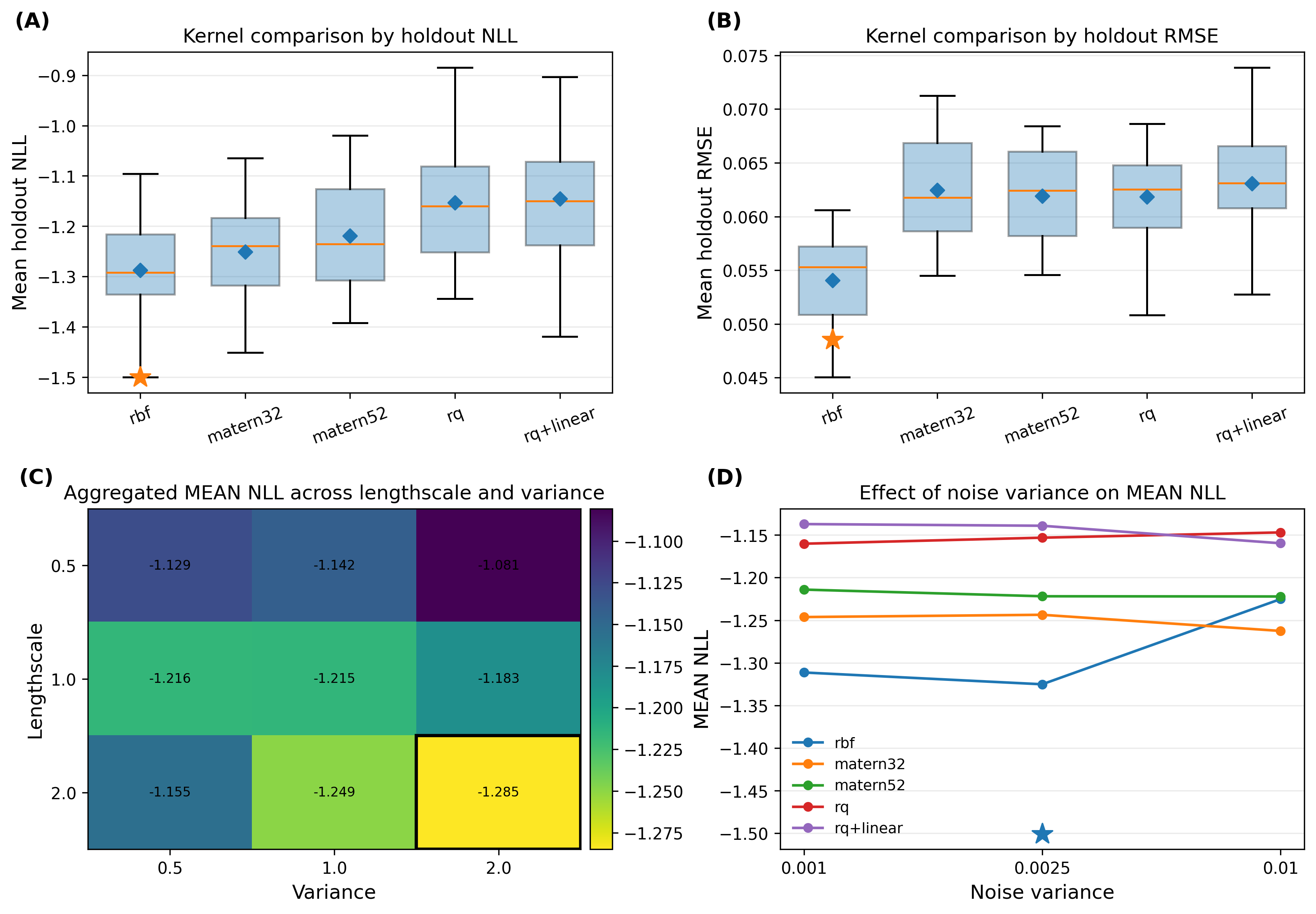}
    \caption{Summary of the Gaussian-process hyperparameter search. 
    (A) Distribution of the mean holdout negative log-likelihood (NLL) across kernel families, showing that the RBF kernel provides the most favorable and most consistent probabilistic generalization among the tested kernels. 
    (B) Distribution of the mean holdout root-mean-square error (RMSE) across kernel families, confirming that the RBF kernel also yields the most accurate point predictions on average. 
    (C) Mean holdout NLL aggregated over the explored lengthscale--variance combinations. The best-performing region is centered around a lengthscale of $2.0$, with variance $2.0$ also lying in a strong-performing regime. 
    (D) Mean holdout NLL as a function of noise variance for each kernel family, indicating that noise variance has a comparatively weaker effect on final performance than kernel choice and lengthscale. 
    The selected configuration, highlighted by the star marker and boxed heatmap cell, corresponds to the RBF kernel with lengthscale $2.0$, variance $2.0$, and noise variance $0.0025$, which achieved the most favorable overall compromise between predictive accuracy and probabilistic calibration within the explored search space.}
    \label{fig:gp_hpo_summary}
\end{figure*}

Figure~\ref{fig:gp_hpo_summary} summarizes the hyperparameter search for the latent-space GP surrogate. The search compared several kernel families and initialization settings for the kernel hyperparameters. Model selection was guided primarily by holdout negative log-likelihood (NLL), while root-mean-square error (RMSE) was also monitored to ensure accurate mean prediction. The latent-space GP surrogate uses covariance functions \(k(\mathbf{u},\mathbf{u}')\) defined on the two-dimensional input
\(\mathbf{u} = [r,n]\), where \(r\) denotes the charging C-rate and \(n\) denotes the cycle number. During hyperparameter
optimization, several kernel families were considered in order to assess the effect of different smoothness assumptions on
the latent trajectories. The candidate kernels are summarized below.

Let \(d(\mathbf{u},\mathbf{u}')\) denote the scaled Euclidean distance,
\[
d(\mathbf{u},\mathbf{u}')^2
=
\sum_{q=1}^{2}\frac{(u_q-u_q')^2}{\ell^2},
\]
where \(\ell\) is the lengthscale hyperparameter. The tested kernels were:

\begin{align}
k_{\mathrm{RBF}}(\mathbf{u},\mathbf{u}')
&=
\sigma_f^2
\exp\!\left(
-\frac{1}{2}d(\mathbf{u},\mathbf{u}')^2
\right),
\\[4pt]
k_{\mathrm{Mat\acute{e}rn-3/2}}(\mathbf{u},\mathbf{u}')
&=
\sigma_f^2
\left(1+\sqrt{3}\,d(\mathbf{u},\mathbf{u}')\right)
\exp\!\left(-\sqrt{3}\,d(\mathbf{u},\mathbf{u}')\right),
\\[4pt]
k_{\mathrm{Mat\acute{e}rn-5/2}}(\mathbf{u},\mathbf{u}')
&=
\sigma_f^2
\left(
1+\sqrt{5}\,d(\mathbf{u},\mathbf{u}')
+\frac{5}{3}d(\mathbf{u},\mathbf{u}')^2
\right)
\exp\!\left(-\sqrt{5}\,d(\mathbf{u},\mathbf{u}')\right),
\\[4pt]
k_{\mathrm{RQ}}(\mathbf{u},\mathbf{u}')
&=
\sigma_f^2
\left(
1+\frac{d(\mathbf{u},\mathbf{u}')^2}{2\alpha}
\right)^{-\alpha},
\\[4pt]
k_{\mathrm{RQ+Linear}}(\mathbf{u},\mathbf{u}')
&=
k_{\mathrm{RQ}}(\mathbf{u},\mathbf{u}')
+\sigma_b^2+\sigma_{\ell}^2\,\mathbf{u}^{\top}\mathbf{u}'.
\end{align}

Here, \(\sigma_f^2\) denotes the output variance, \(\alpha\) is the rational-quadratic shape parameter, and the linear term
allows the GP to represent an additional global linear trend on top of the multi-scale smooth component captured by the
RQ kernel.

Overall, the RBF kernel provided the most favorable trade-off between predictive accuracy and probabilistic calibration. In addition, the most favorable region of the explored search space was centered around a lengthscale of $2.0$, with variance $2.0$ also lying in a strong-performing regime. By contrast, the effect of the noise variance was comparatively weak across the explored values (see Figure~\ref{fig:gp_hpo_summary}).

\begin{table}[!t]
\centering
\caption{Selected kernel and training settings for the 2D Gaussian-process latent surrogate. The GP takes cycle number and charging C-rate as inputs and predicts the two latent coordinates $(z_1,z_2)$.}
\label{tab:si_gp_hparams}
\small
\begin{tabular}{p{0.42\linewidth} p{0.46\linewidth}}
\toprule
\textbf{Setting} & \textbf{Value} \\
\midrule
Kernel & RBF \\
Selected lengthscale & 2.0 \\
Selected output variance & 2.0 \\
Selected noise variance & 0.0025 \\
Number of inducing points & 256 \\
Batch size & 512 \\
Evaluation batch size & 8192 \\
Optimizer & Adam \\
Learning rate & 0.01 \\
Scheduler & ReduceLROnPlateau \\
Scheduler factor & 0.5 \\
Scheduler patience & 3 \\
Minimum learning rate & \(10^{-5}\) \\
Early stopping monitor & Validation NLL \\
Early stopping patience & 8 \\
Early stopping min\_delta & \(10^{-4}\) \\
Model-selection metric & NLL \\
\bottomrule
\end{tabular}
\end{table}

The following figure~\ref{fig:si_gp_test_uncertainty} illustrates the predictive uncertainty of the GP surrogate on the two outer-test charging trajectories.

\begin{figure*}[!ht]
  \centering
  \includegraphics[width=0.8\textwidth,keepaspectratio]{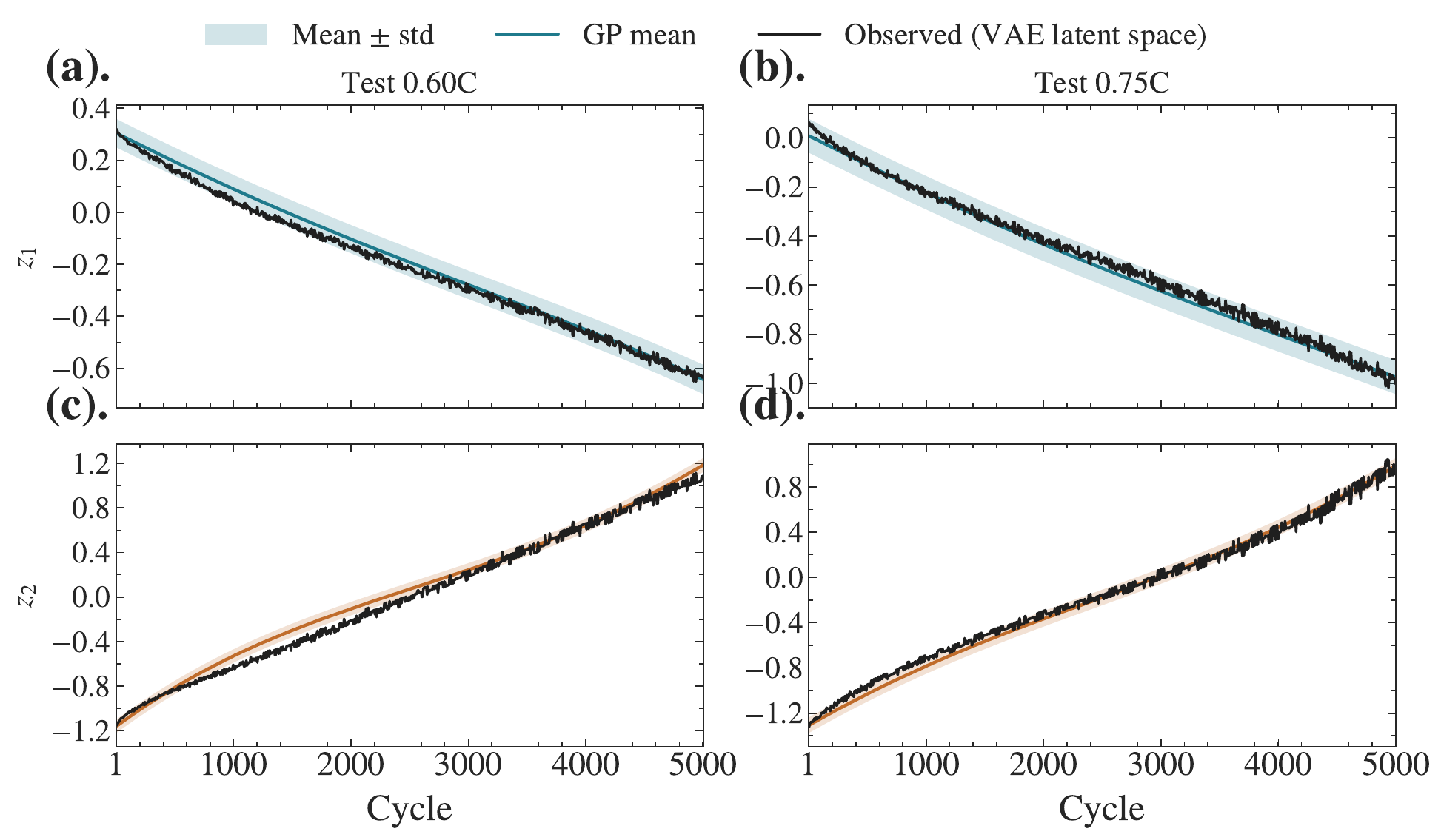}
  \caption{Cycle-wise GP predictions of the latent trajectories for the outer-test charging rates. Panels (a) and (b) show the GP prediction for latent coordinate $z_1$ at outer-test 0.60C and 0.75C, respectively, while panels (c) and (d) show the corresponding predictions for $z_2$. In each panel, the black curve denotes the reference latent trajectory obtained from the trained VAE encoder, the colored solid curve denotes the GP posterior mean, and the shaded band indicates the predictive uncertainty. The close agreement between the GP posterior mean and the reference trajectory indicates that the GP captures the smooth latent evolution over cycle life, while the uncertainty bands reflect reduced confidence in regions where support from neighboring charging trajectories is weaker.}
  \label{fig:si_gp_test_uncertainty}
\end{figure*}

\clearpage
\section{Decoding Votlage channels and SOH values}

\begin{algorithm}[H]
\caption{Predicting voltage channels and SOH using GP posterior and frozen decoder}
\label{alg:gp_decoder_inference}
\begin{algorithmic}[1]
\Require Trained latent GP models $\{\mathcal{GP}_j\}_{j=1}^d$, frozen decoder $D_{\theta^\star}$, optional SOH head $g_{\psi^\star}$, query condition $(r^\ast,n^\ast)$, capacity grid $\mathbf{q}$, number of latent samples $S$
\Ensure Predictive voltage channels and SOH statistics at $(r^\ast,n^\ast)$

\For{latent dimension $j=1$ to $d$}
    \State Query GP posterior at $\mathbf{u}^\ast=[r^\ast,n^\ast]$
    \[
    z_j^\ast \sim \mathcal{N}(\mu_j^\ast,\,\sigma_j^{\ast 2})
    \]
\EndFor

\State Construct latent predictive distribution
\[
p(\mathbf{z}^\ast\mid r^\ast,n^\ast)
=
\mathcal{N}(\boldsymbol{\mu}^\ast,\mathbf{\Sigma}^\ast)
\]
where $\boldsymbol{\mu}^\ast=[\mu_1^\ast,\dots,\mu_d^\ast]$ and $\mathbf{\Sigma}^\ast$ is the latent covariance
\State Freeze decoder parameters $\theta^\star$

\For{$s=1$ to $S$}
    \State Sample latent code:
    \[
    \mathbf{z}^{\ast(s)} \sim \mathcal{N}(\boldsymbol{\mu}^\ast,\mathbf{\Sigma}^\ast)
    \]
    \State Decode voltage-related channels:
    \[
    \hat{\mathbf{x}}^{(s)} \gets D_{\theta^\star}(\mathbf{z}^{\ast(s)},\mathbf{q})
    \]
    \State Extract desired outputs, e.g.
    \[
    \hat{V}_{\mathrm{ch}}^{(s)}(\mathbf{q}),\quad
    \hat{V}_{\mathrm{dis}}^{(s)}(\mathbf{q})
    \]
    \If{SOH head is used}
        \State Predict SOH from sampled latent:
        \[
        \hat{y}_{\mathrm{SOH}}^{(s)} \gets g_{\psi^\star}(\mathbf{z}^{\ast(s)})
        \]
    \Else
        \State Compute SOH from decoded channel or reconstructed capacity proxy, if available
    \EndIf
\EndFor

\State Estimate predictive mean and uncertainty of voltage channels:
\[
\bar{V}_{\mathrm{ch}}(\mathbf{q})
=
\frac{1}{S}\sum_{s=1}^S \hat{V}_{\mathrm{ch}}^{(s)}(\mathbf{q}),
\qquad
\bar{V}_{\mathrm{dis}}(\mathbf{q})
=
\frac{1}{S}\sum_{s=1}^S \hat{V}_{\mathrm{dis}}^{(s)}(\mathbf{q})
\]

\State Estimate predictive SOH mean and uncertainty:
\[
\bar{y}_{\mathrm{SOH}}
=
\frac{1}{S}\sum_{s=1}^S \hat{y}_{\mathrm{SOH}}^{(s)}
\]
\[
\mathrm{Var}(y_{\mathrm{SOH}})
=
\frac{1}{S-1}\sum_{s=1}^S
\left(\hat{y}_{\mathrm{SOH}}^{(s)}-\bar{y}_{\mathrm{SOH}}\right)^2
\]

\State Return decoded voltage channels, SOH prediction, and predictive intervals
\end{algorithmic}
\end{algorithm}

%% file: affliations.tex
\title{BattVAE-GP: Generative Modeling of Long-Horizon \\ Battery Degradation with Uncertainty Quantification}


\author[lrcs,rs2e,predeeption]{Raghvender Raghvender}

\author[lrcs,rs2e]{Mahdi Abid}

\author[warwick,faraday]{Ferran Brosa Planella}

\author[lrcs,rs2e]{Charles Delacourt}

\author[lrcs,rs2e,predeeption]{Arnaud Demortière\corref{cor1}}
\ead{arnaud.demortiere@cnrs.fr}

\cortext[cor1]{Corresponding author.}

\affiliation[lrcs]{organization={Laboratoire de Réactivité et de Chimie des Solides (LRCS), UMR CNRS 7314-Université de Picardie Jules Verne},
            addressline={15 Rue Baudelocque}, 
            postcode={80000},
            city={Amiens},
            country={France}}

\affiliation[rs2e]{organization={Réseau sur le Stockage Electrochimique de l’Energie (RS2E), CNRS FR 3459, Hub de l’Energie},
            addressline={15 Rue Baudelocque}, 
            postcode={80000},
            city={Amiens},
            country={France}}

\affiliation[predeeption]{organization={Predeeption (INRIA Startup Studio / CNRS Innovation)},
            addressline={Hub de l'energie},
            postcode={80000},
            city={Amiens},
            country={France}}

\affiliation[warwick]{organization={Mathematics Institute},
            addressline={University of Warwick, Gibbet Hill Road}, 
            postcode={CV4 7AL},
            city={Coventry},
            country={UK}}

\affiliation[faraday]{organization={The Faraday Institution},
            addressline={Quad One, Becquerel Avenue, Harwell Campus}, 
            postcode={OX11 0RA},
            city={Didcot},
            country={UK}}